\documentclass{article}

\usepackage{iclr2021_conference,times}
\iclrfinalcopy
 \usepackage[printonlyused]{acronym}
\usepackage[export]{adjustbox} %
\usepackage{algorithm}
\usepackage{algorithmicx}
\usepackage{algpseudocode}
\usepackage{amssymb}
\usepackage{amsmath}
\usepackage{array}
\usepackage{bbm}
\usepackage{booktabs}
\usepackage{caption}
\usepackage{float}
\usepackage{graphicx}
\usepackage{hyperref}
\usepackage{lipsum}
\usepackage{microtype}
\usepackage{multicol}
\usepackage{multirow}
\usepackage{natbib}
\usepackage{nicefrac}
\usepackage{relsize}
\usepackage{siunitx}
\usepackage{subcaption}
\usepackage{wrapfig}
\usepackage{tikz}
\usetikzlibrary{shapes, arrows, positioning, fit, backgrounds}
\usetikzlibrary{calc}
\usetikzlibrary{decorations.pathmorphing}
\tikzstyle{block} = [rectangle, draw, fill=blue!20,
    text centered, rounded corners, minimum height=1em, node distance=1.5cm]
\tikzstyle{line} = [draw, -latex']
\tikzstyle{cloud} = [draw, rectangle,fill=red!20, node distance=1.5cm and 1cm,
    minimum height=1em]
\usepackage{verbatim}
\usepackage{xspace}

\usepackage{amsthm}
\algnewcommand{\IfThenElse}[3]{%
  \State \algorithmicif\ #1\ \algorithmicthen\ #2\ \algorithmicelse\ #3}
 \algnewcommand{\IfThen}[2]{%
  \State \algorithmicif\ #1\ \algorithmicthen\ #2}

\newtheorem*{theorem-nonlabeled}{Theorem}

\usepackage[utf8]{inputenc} %
\usepackage[T1]{fontenc}    %
\usepackage{hyperref}       %
\usepackage{url}            %
\usepackage{booktabs}       %
\usepackage{amsfonts}       %
\usepackage{nicefrac}       %
\usepackage{microtype}      %

\definecolor{mydarkblue}{rgb}{0,0.08,0.45}
\hypersetup{ %
    pdftitle={},
    pdfauthor={},
    pdfsubject={},
    pdfkeywords={},
    pdfborder=0 0 0,
    pdfpagemode=UseNone,
    colorlinks=true,
    linkcolor=mydarkblue,
    citecolor=mydarkblue,
    filecolor=mydarkblue,
    urlcolor=mydarkblue,
    pdfview=FitH}

\algnewcommand{\algorithmicforeach}{\textbf{for each}}
\algdef{SE}[FOR]{ForEach}{EndForEach}[1]
  {\algorithmicforeach\ #1\ \algorithmicdo}%
  {\algorithmicend\ \algorithmicforeach}%
\newcommand{\pushright}[1]{\ifmeasuring@#1\else\omit\hfill$\displaystyle#1$\fi\ignorespaces}

\newcommand{\data}{\mathcal{D}}

\newcommand{\upd}{\mathcal{U}}

\newcommand{\bz}{\mathbf{z}}

\newcommand{\bs}{\mathbf{s}}

\newcommand{\ent}{\mathcal{H}}

\newcommand{\changes}[1]{#1}

\newcommand{\refEquation}[1]{Eq. \ref{#1}}
\newcommand{\refSection}[1]{Section \ref{#1}}
\newcommand{\refFigure}[1]{Figure~\ref{#1}}
\newcommand{\refTable}[1]{Table~\ref{#1}}
\newcommand{\refAlgorithm}[1]{Algorithm~\ref{#1}}

\newcommand{\valueWithUnits}[2]{#1~\normalsize #2\normalsize}

\DeclareMathOperator*{\argmax}{arg\,max \:}

\newcommand{\RL}{\ac{RL}\xspace}

\newcommand{\methodName}{SMiRL\xspace}
\newcommand{\entropic}{unstable\xspace}

\newcommand{\Tetris}{\textit{Tetris}\xspace}
\newcommand{\VizDoom}{\textit{VizDoom}\xspace}
\newcommand{\VizDoomTakeCover}{\textit{TakeCover}\xspace}
\newcommand{\VizDoomDefendTheLine}{\textit{DefendTheLine}\xspace}
\newcommand{\humanoid}{\textit{Humanoid}\xspace}
\newcommand{\humanoidCliff}{\textit{Cliff}\xspace}
\newcommand{\humanoidTreadmill}{\textit{Treadmill}\xspace}
\newcommand{\humanoidWalk}{\textit{Walk}\xspace}

\newcommand{\humanoidPedestal}{\textit{Pedestal}\xspace}
\newcommand{\miniGrid}{\textit{HauntedHouse}\xspace}

\def\signed #1{{\leavevmode\unskip\nobreak\hfil\penalty50\hskip2em
  \hbox{}\nobreak\hfil(#1)%
  \parfillskip=0pt \finalhyphendemerits=0 \endgraf}}
\newsavebox\mybox

\title{SMiRL: Surprise Minimizing Reinforcement Learning in Unstable Environments}

\author{  Glen Berseth \\
  UC Berkeley\\
  \And
  Daniel Geng \\
  UC Berkeley\\
  \And
   Coline Devin \\
   UC Berkeley\\
 \And
 Nicholas Rhinehart \\
 UC Berkeley\\
 \AND
 Chelsea Finn \\
 Stanford
 \And
 Dinesh Jayaraman \\
 UPenn\\
 \And
 Sergey Levine \\
 UC Berkeley\\
  }
\begin{document}
\maketitle

\acrodef{AGI}{artificial general intelligence}
\acrodef{ANOVA}[ANOVA]{Analysis of Variance\acroextra{, a set of
  statistical techniques to identify sources of variability between groups}}
\acrodef{ANN}{artificial neural network}
\acrodef{API}{application programming interface}
\acrodef{CACLA}{continuous actor critic learning automaton}
\acrodef{cGAN}{conditional generative adversarial network}
\acrodef{CMA}{covariance matrix adaptation}
\acrodef{COM}{centre of mass}
\acrodef{CTAN}{\acroextra{The }Common \TeX\ Archive Network}
\acrodef{DDPG}{deep deterministic policy gradient}
\acrodef{DeepLoco}{deep locomotion}
\acrodef{DOI}{Document Object Identifier\acroextra{ (see
    \url{http://doi.org})}}
\acrodef{DPG}{deterministic policy gradient}
\acrodef{DQN}{deep Q-network}
\acrodef{DRL}{deep reinforcement learning}
\acrodef{DYNA}{DYNA}
\acrodef{EOM}{Equations of motion}
\acrodef{EPG}{expected policy gradient}
\acrodef{FDR}{future discounted reward}
\acrodef{FSM}{finite state machine}
\acrodef{GAE}{generalized advantage estimation}
\acrodef{GAN}{generative adversarial network}
\acrodef{GPS}[GPS]{Graduate and Postdoctoral Studies}
\acrodef{HLC}{high-level controller}
\acrodef{HLP}{high-level policy}
\acrodef{HRL}{hierarchical reinforcement learning}
\acrodef{KLD}{Kullback-Leibler divergence}
\acrodef{LLC}{low-level controller}
\acrodef{LLP}{low-level policy}
\acrodef{MARL}{Multi-Agent Reinforcement Learning}
\acrodef{MBAE}{model-based action exploration}
\acrodef{MPC}{model predictive control}
\acrodef{MDP}{Markov Decision Processes}
\acrodef{MSE}{mean squared error}
\acrodef{MultiTasker}{controller that learns multiple tasks at the same time}
\acrodef{Parallel}{randomly initialize controllers and train in parallel}
\acrodef{PD}{proportional derivative}
\acrodef{PDF}{Portable Document Format}
\acrodef{PLAiD}{Progressive Learning and Integration via Distillation}
\acrodef{PPO}{proximal policy optimization}
\acrodef{PTD}{positive temporal difference}
\acrodef{RBF}{radial basis function}
\acrodef{ReLU}{rectified linear unit}
\acrodef{RCS}[RCS]{Revision control system\acroextra{, a software
    tool for tracking changes to a set of files}}
\acrodef{RL}{reinforcement learning}
\acrodef{SGD}{stochastic gradient descent}
\acrodef{Scratch}{randomly initialized controller}
\acrodef{SIMBICON}{SIMple BIped CONtroller}
\acrodef{SMBAE}{stochastic model-based action exploration}
\acrodef{SVG}{stochastic value gradients}
\acrodef{SVM}{support vector machine}
\acrodef{TL}{transfer learning}
\acrodef{TD}{temporal difference}
\acrodef{terrainRL}{terrain adaptive locomotion}
\acrodef{TLX}[TLX]{Task Load Index\acroextra{, an instrument for gauging
  the subjective mental workload experienced by a human in performing
  a task}}
\acrodef{TRPO}{trust region policy optimization}
\acrodef{UBC}{University of British Columbia}
\acrodef{UCB}{upper confidence bound}
\acrodef{UI}{user interface}
\acrodef{UML}{Unified Modelling Language\acroextra{, a visual language
    for modelling the structure of software artefacts}}
\acrodef{URDF}{unified robot description format}
\acrodef{URL}{Unique Resource Locator\acroextra{, used to describe a
    means for obtaining some resource on the world wide web}}
\acrodef{W3C}[W3C]{\acroextra{the }World Wide Web Consortium\acroextra{,
    the standards body for web technologies}}    
\acrodef{XML}{Extensible Markup Language}

\setlength\abovecaptionskip{0.1cm}

\begin{abstract}
Every living organism struggles against disruptive environmental forces to carve out and maintain an orderly niche. We propose that such a struggle to achieve and preserve order might offer a principle for the emergence of useful behaviors in artificial agents. We formalize this idea into an unsupervised reinforcement learning method called surprise minimizing reinforcement learning (SMiRL). SMiRL alternates between learning a density model to evaluate the surprise of a stimulus, and improving the policy to seek more predictable stimuli. The policy seeks out stable and repeatable situations that counteract the environment's prevailing sources of entropy. This might include avoiding other hostile agents, or finding a stable, balanced pose for a bipedal robot in the face of disturbance forces. We demonstrate that our surprise minimizing agents can successfully play Tetris, Doom, control a humanoid to avoid falls, and navigate to escape enemies in a maze without any task-specific reward supervision. We further show that SMiRL can be used together with standard task rewards to accelerate reward-driven learning.
\end{abstract}
\section{Introduction}
\label{sec:intro}

Organisms can carve out environmental niches within which they can maintain relative predictability amidst the entropy around them~\citep{boltzmann1886second,schrodinger1944life,schneider1994life,friston2009free}. For example, humans go to great lengths to shield themselves from surprise --- we band together to build cities with homes, supplying water, food, gas, and electricity to control the deterioration of our bodies and living spaces amidst heat, cold, wind and storm. These activities exercise sophisticated control over the environment, which makes the environment more predictable and less ``surprising'' \citep{friston2009free,10.1371/journal.pone.0006421}. Could the motive of preserving order guide the automatic acquisition of useful behaviors in artificial agents?

We study this question in the context of unsupervised reinforcement learning, which deals with the problem of acquiring complex behaviors and skills with no supervision (labels) or incentives (external rewards). Many previously proposed unsupervised reinforcement learning methods focus on novelty-seeking behaviors~\citep{schmidhuber1991curious,lehman2011abandoning,still2012information,bellemare2016unifying,Houthooft2016,Pathak2017}.
Such methods can lead to meaningful behavior in simulated environments, such as video games, where interesting and novel events mainly happen when the agent executes a specific and coherent pattern of behavior. However, we posit that in more realistic open-world environments, natural forces outside of the agent's control \emph{already} offer an excellent source of novelty: from other agents to unexpected natural forces, agents in these settings must contend with a constant stream of unexpected events. In such settings, rejecting perturbations and maintaining a steady equilibrium may pose a greater challenge than novelty seeking. Based on this observation, we devise an algorithm, surprise minimizing reinforcement learning (\methodName), that specifically aims to \emph{reduce} the entropy of the states visited by the agent.

\begin{figure}[tb]
    \centering
    \subcaptionbox{\label{fig:smirl} }{
    \includegraphics[width=0.45\textwidth]{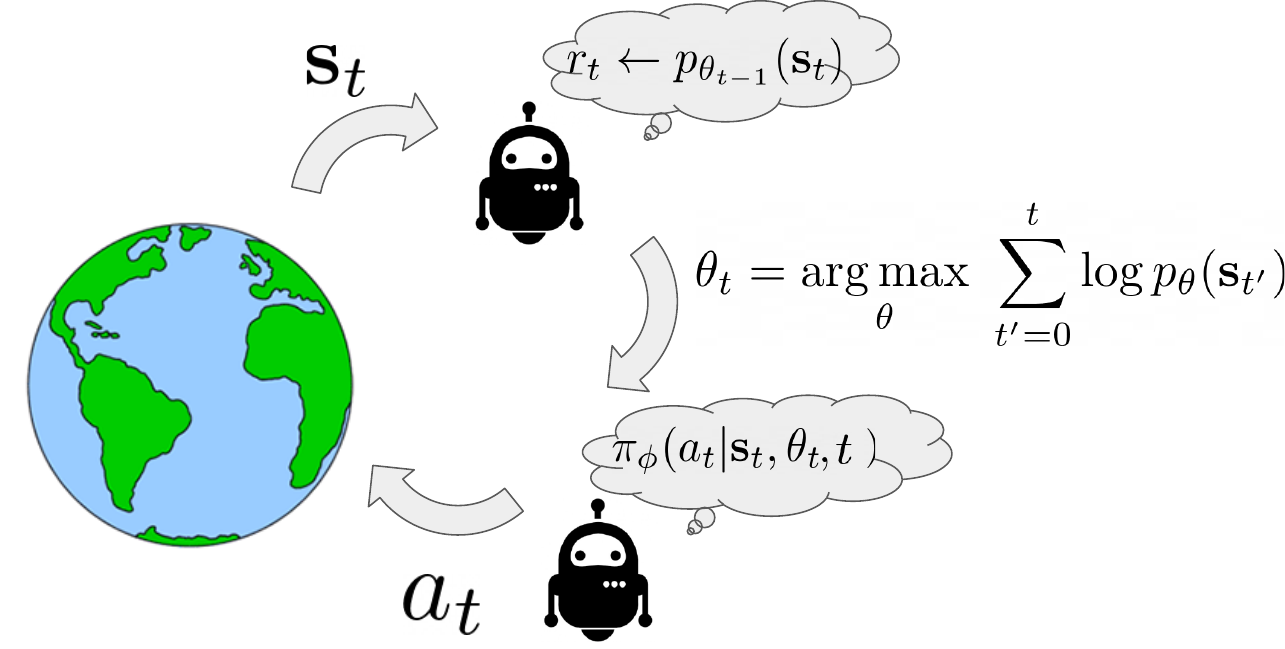}}
    \subcaptionbox{\label{fig:teaser} }{\includegraphics[width=0.45\textwidth]{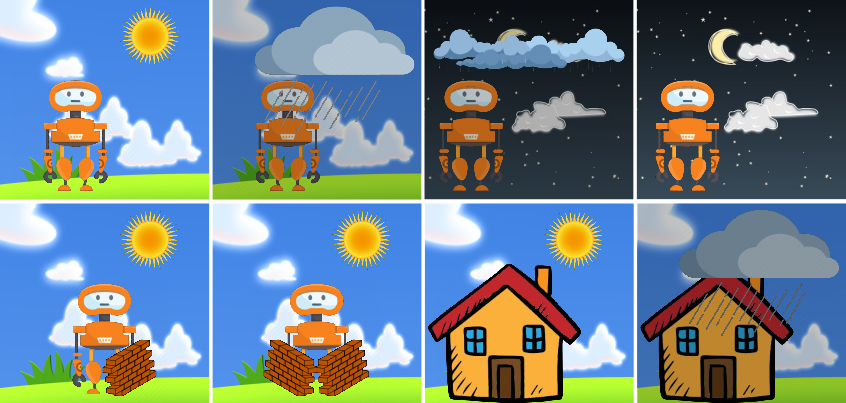}}
    \caption{{\footnotesize \textbf{Left:} \methodName observes a state $\bs_t$ and computes a reward $r_{t}$ as the negative \textit{surprise} under its current model $p_{\theta_{t-1}}(\bs_t)$, given by $\log p_{\theta_{t-1}}(\bs_t)$. Then the model is updated on the agents state history, including $\bs_t$, to yield $p_{\theta_{t}}$. The policy $\pi_{\phi}(a_t|\bs_t, \theta_{t}, t)$ then generates the action $a_{t}$. \textbf{Right:} This procedure leads to complex behavior in environments where surprising events happen on their own. In this cartoon, the robot experiences a wide variety of weather conditions when standing outside, but can avoid these surprising conditions by building a shelter, where it can reach a stable and predictable states in the long run.
    }
    \vspace{-0.2in}
    }
\end{figure}

\methodName maintains an estimate of the distribution of visited states, $p_{\theta}(\bs)$,
and a policy that seeks to reach likely future states under $p_{\theta}(\bs)$. 
After each action, $p_{\theta}(\bs)$ is updated with the new state, while the policy is conditioned on the parameters of this distribution to construct a stationary MDP. We illustrate this with a diagram in~\refFigure{fig:smirl}.
We empirically evaluate \methodName in a range of domains that are characterized by naturally increasing entropy, including video game environments based on Tetris and Doom, and simulated robot tasks that require controlling a humanoid robot to balance and walk. Our experiments show that, in environments that satisfy the assumptions of our method, \methodName automatically discovers complex and coordinated behaviors without any reward signal, learning to successfully play Tetris, shoot enemies in Doom, and balance a humanoid robot at the edge of a cliff. We also show that \methodName can provide an effective auxiliary objective when a reward signal is provided, accelerating learning in these domains substantially more effectively than pure novelty-seeking methods.
Videos of our results are available online\footnote{https://sites.google.com/view/surpriseminimization}

\section{Related Work}

Prior work on unsupervised learning has proposed algorithms that learn without a reward function, such as empowerment~\citep{10.1007/11553090_75,NIPS2015_5668} or intrinsic motivation~\citep{chentanez2005intrinsically,oudeyer2009intrinsic,oudeyer2007intrinsic}.
Intrinsic motivation has typically focused on encouraging novelty-seeking behaviors by maximizing model uncertainty~\citep{Houthooft2016,still2012information,shyam2018model,Pathak2019}, by maximizing model prediction error or improvement~\citep{lopes2012exploration,Pathak2017}, through state visitation counts~\citep{bellemare2016unifying},  via surprise maximization~\citep{achiam2017surprise,schmidhuber1991curious,sun2011planning}, and through other novelty-based reward bonuses~\citep{lehman2011abandoning,DBLP:journals/corr/AchiamS17,Burda2018,kim2019curiosity}. We do the opposite. Inspired by the free energy principle~\citep{friston2009free,10.1371/journal.pone.0006421,ueltzhoffer2018deep,faraji2018balancing,friston2016active} \changes{including recent methods that train policies using RL~\citep{tschantz2020scaling,tschantz2020reinforcement,annabi2020autonomous} that encode a prior over desired observations}, 
we instead incentivize an agent to \emph{minimize} surprise \changes{over the distribution of states generated by the policy in unstable environments}, and study the resulting behaviors. In such environments \changes{it is non-trivial to achieve low entropy state distributions}, which we believe are more reflective of the real world. \changes{Learning progress methods that minimize model parameter entropy~\citep{lopes2012exploration,kim2020active} avoid the issues novelty-based methods have with noisy distractors. These methods are based on learning the parameters of the dynamics where our method is learning to control the marginal state distribution.}

Several works aim to maximize state entropy to encourage exploration~\citep{lee2019efficient,hazan2019provably}. Our method aims to do the opposite, \emph{minimizing} state entropy. \changes{Recent work connects the free energy principle, empowerment and predictive information maximization under the same framework to understand their differences~\citep{Biehl2018FreeE}.}
Existing work has also studied how competitive self-play and competitive, multi-agent environments can lead to complex behaviors with minimal reward information~\citep{silver2017mastering,bansal2017emergent,sukhbaatar2017intrinsic,baker2019emergent,weihs2019artificial,chen2020visual}. Like these works, we also consider how complex behaviors can emerge in resource-constrained environments, but instead of multi-agent competition, we utilize surprise minimization to drive the emergence of complex skills.

\section{Surprise Minimizing Agents}
\label{sec:method}

We propose surprise minimization as a means to operationalize the idea of learning useful behaviors by seeking out low entropy state distributions. 
The long term effects of actions on surprise can be subtle, since actions change both (i) the state that the agent is in, and (ii) its beliefs, represented by a model $p_{\theta}(\bs)$, about which states are likely under its current policy.
\methodName induces the agent to modify its policy $\pi$ so that it encounters states $\bs$ with high $p_{\theta}(\bs)$, as well as to seek out states that will change the model $p_{\theta}(\bs)$ so that future states are more likely.
In this section, we will first describe what we mean by \entropic environments and provide the surprise minimization problem statement, and then present our practical deep reinforcement learning algorithm for learning policies that minimize surprise.

Many commonly used reinforcement learning benchmark environments are \emph{stable}, in the sense the agent remains in a narrow range of starting states unless it takes coordinated and purposeful actions. In such settings, unsupervised RL algorithms that seek out novelty can discover meaningful behaviors. However, many environments -- including, as we argue, those that reflect properties commonly found in the real world, -- are \entropic, in the sense that unexpected and disruptive events naturally lead to novelty and increased state entropy even if the agent does not carry out any particularly meaningful or purposeful behavior. 
In \entropic environments, minimizing cumulative surprise requires taking actions to reach a stable distribution of states, and then acting continually and purposefully to stay in this distribution.
An example of this is illustrated in~\refFigure{fig:teaser}: the agent's environment is \entropic due to varied weather. If the robot builds a shelter, it will initially experience unfamiliar states, but in the long term the observations inside the shelter are more stable and less surprising than those outside. Another example is the game of Tetris (\refFigure{fig:discreteenvs}),
where the environment spawns new blocks and drops them into random configurations, unless a skilled agent takes actions to control the board. 
The challenge of maintaining low entropy in \entropic settings forces the \methodName agent to acquire meaningful skills.
We defer a more precise definition of \entropic environments to \refSection{sec:dynamic-envs}, where we describe several \entropic environments and contrast them with the static environments that are more commonly found in RL benchmark tasks. In static environments, novelty seeking methods must discover complex behaviors to increase entropy, leading to interesting behavior, while \methodName may trivially find low entropy policies. We show that the reverse is true for \entropic environments: a novelty seeking agent is satisfied with watching the environment change around it, while a surprise minimizing agent must develop meaningful skills to lower entropy.

\paragraph{Problem statement.}
To instantiate \methodName, we design a reinforcement learning agent that receives larger rewards for experiencing more familiar states, based on the history of states it has experienced during the current episode. This translates to learning a policy with the lowest state entropy. We assume a fully-observed controlled Markov process (CMP), where we use $\bs_t$ to denote the state at time $t$, $a_t$ to denote the agent's action, $p(\bs_0)$ to denote the initial state distribution, and $T(\bs_{t+1}|\bs_t, a_t)$ to denote the transition probabilities. The agent learns a policy $\pi_\phi(a|\bs)$, parameterized by $\phi$.
The goal is to minimize the entropy of its state marginal distribution under its current policy $\pi_\phi$ at each time step of the episode. We can estimate this entropy by fitting an estimate of the state marginal $d^{\pi_\phi}(\bs_t)$ at each time step $t$, given by $p_{\theta_{t-1}}(\bs_t)$, using the states seen so far during the episode, $\tau_t = \{ \bs_1, \dots, \bs_t \}$ that is stationary. The sum of the entropies of the state distributions over an episode can then be estimated as
\begin{equation}
\label{eq:smirl-ent-approx}
\sum_{t=0}^T \ent(\bs_t) = -\sum_{t=0}^T \mathbb E_{\bs_t \sim d^{\pi_\phi}(\bs_t)}[\log d^{\pi_\phi}(\bs_t)] \leq -\sum_{t=0}^T \mathbb E_{\bs_t \sim d^{\pi_\phi}(\bs_t)}[\log p_{\theta_{t-1}}(\bs_t)], %
\end{equation}
where the inequality becomes an equality if $p_{\theta_{t-1}}(\bs_t)$ accurately models $d^{\pi_\phi}(\bs_t)$. Minimizing the right-hand side of this equation corresponds to maximizing an RL objective with rewards:
\begin{equation}
r(\bs_t) = \log p_{\theta_{t-1}}(\bs_t). \label{eq:smirl_reward}
\end{equation}
However, an optimal policy for solving this problem must take changes in the distribution $p_{\theta_{t-1}}(\bs_t)$ into account when selecting actions, since this distribution changes at each step. To ensure that the underlying RL optimization corresponds to a stationary and Markovian problem, we construct an \emph{augmented} MDP to instantiate \methodName in practice, which we describe in the following section.

\paragraph{Training \methodName agents.}%

\begin{wrapfigure}{R}{0.5\textwidth}
\vspace{-0.25cm}
\begin{minipage}{.5\textwidth}
\begin{algorithm}[H]
\footnotesize 	
\caption{\methodName}
\label{alg:training}
\begin{algorithmic}[1]
\While{not converged}
\State $\beta\gets \{\}$ \algorithmiccomment{Reset experience}
\For{$\text{episode } = 0,\dots,M$}
\State $\bs_0 \sim p(\bs_0); \tau_0 \gets  \{\bs_0\}$ \algorithmiccomment{Initialize state}
\State $\bar{\bs}_0 \gets (\bs_0, \mathbf{0}, 0)$ \algorithmiccomment{Initialize aug. state}
\ForEach{$t = 0,\dots,T$}
\State $a_{t} \sim \pi_\phi(a_t| \bs_{t}, \theta_{t}, t)$\algorithmiccomment{Get  action}
\State $\bs_{t+1} \sim T(\bs_{t+1} | \bs_{t}, a_{t})$  \algorithmiccomment{Step dynamics}
\State $r_{t} \gets  \log p_{\theta_{t}}(\bs_{t+1})$ \algorithmiccomment{\methodName reward}
\State $\tau_{t+1}\!\gets\!\tau_t\cup \{\bs_{t+1}\}$ \algorithmiccomment{Record state}
\State $\theta_{t+1} \gets \upd(\tau_{t+1})$ \algorithmiccomment{Fit model}
\State $\bar{\bs}_{t+1} \gets \{( \bs_{t+1}, \theta_{t+1}, t_{t+1})\}$
\State $\beta \gets \beta \cup \{(\bar{\bs}_{t}, a_t, r_t, \bar{\bs}_{t+1})\}$
\EndFor
\EndForEach
\State $\phi \gets \texttt{RL} (\phi, \beta)$ \algorithmiccomment{Update policy}

\EndWhile
\end{algorithmic}
\end{algorithm}
\end{minipage}
\vspace{-0.25cm}
\end{wrapfigure}
In order to instantiate \methodName, we construct an \emph{augmented} MDP out of the original CMP, where the reward in Equation~(\ref{eq:smirl_reward}) can be expressed entirely as a function of the state. This augmented MDP has a state space that includes the original state $\bs_t$, as well as the sufficient statistics of $p_{\theta_t}(\bs)$. For example, if $p_{\theta_t}(\bs)$ is a normal distribution with parameters ${\theta_t}$, then $({\theta_t},t)$ -- the parameters of the distribution and the number of states seen so far -- represents a sufficient statistic. 
Note that it is possible to use other, more complicated, methods to summarize the statistics, including reading in the entirety of $\tau_t$ using a recurrent model. 
The policy conditioned on the augmented state is then given by $\pi_\phi(a_t|\bs_t,\theta_t,t)$. The parameters of the sufficient statistics are updated $\theta_{t} \!=\!\upd(\tau_{t})$ using a maximum likelihood state density estimation process \mbox{$\theta_{t} \!=\! \argmax_{\theta} \sum_{n=0}^t \log p_{\theta}(\bs_{n})$} over the experience within the episode $\tau_{t}$.
When $(\theta_t, t)$ is a sufficient statistic, the update may be written as $\theta_{t} = \upd(\bs_{t}, \theta_{t-1}, t-1)$.
Specific update functions $\upd(\tau_{t})$ used in our experiments are described in Appendix~\ref{sec:smirl_distributions} and at the end of the section.
Since the reward is given by $r(\bs_t,\theta_{t-1},t-1) = \log p_{\theta_{t-1}}(\bs_t)$, and $\theta_{t}$ is a function of $\bs_t$ and $(\theta_{t-1}, t-1)$, the resulting RL problem is fully Markovian and stationary, and as a result standard RL algorithms will converge to locally optimal solutions. Appendix~\ref{app:smirl_MDP} include details on the MDP dynamics.
In~\refFigure{fig:smirl_rollout}, we illustrate the evolution of $p_{\theta_t}(\bs)$ during an episode of the game Tetris.
The pseudocode for this algorithm is presented in~\refAlgorithm{alg:training}. 

\paragraph{Density estimation with learned representations.}
\methodName may, in principle, be used with any choice of model class for the density model $p_{\theta_t}(\bs)$. As we show in our experiments, relatively simple distribution classes, such as products of independent marginals, suffice to run \methodName in many environments. However, it may be desirable in more complex environments to use more sophisticated density estimators, especially when learning directly from high-dimensional observations such as images. In these cases, we can use variational autoencoders (VAEs)~\citep{kingma2013auto} to learn a non-linear state representation. A VAE is trained using the standard ELBO objective to reconstruct states $\bs$ after encoding them into a latent representation $\bz$ via an encoder $q_\omega(\bz|\bs)$, with parameters $\omega$. Thus, $\bz$ can be viewed as a compressed representation of the state.

When using VAE representations, we train the VAE online together with the policy. This approach necessitates two changes to the procedure described Algorithm~\ref{alg:training}.
First, training a VAE requires more data than the simpler independent models, which can easily be fitted to data from individual episodes. We propose to overcome this by not resetting the VAE parameters between training episodes, and instead training the VAE across episodes.
Second, instead of passing the VAE model parameters to the policy, we only update a distribution over the VAE latent state, given by $p_{\theta_{t}}(\bz)$, such that $p_{\theta_{t}}(\bz)$ replaces $p_{\theta_t}(\bs)$ in the \methodName algorithm, and is fitted to only that episode's (encoded) state history. We represent $p_{\theta_{t}}(\bz)$ as a normal distribution with a diagonal covariance, and fit it to the VAE encoder outputs.
Thus, the mean and variance of $p_{\theta_{t}}(\bz)$ are passed to the policy at each time step, along with $t$.
This implements the density estimate in line 9 of Algorithm~\ref{alg:training}. 
The corresponding update $\upd(\tau_t)$ is:
\begin{align*}
    &\bz_0, \dots, \bz_t = \mathbb{E}[q_\omega(\bz|\bs)] \text{ for } \bs \in \tau_t, 
    &\mu = \nicefrac{1}{t+1}{\sum_{j=0}^t \bz_j}, \sigma = \nicefrac{1}{t+1}{\sum_{j=0}^t (\mu-\bz_j)^2}{}, 
    &\theta_t = [\mu, \sigma].
\end{align*}
Training the VAE online, over all previously seen data, deviates from the recipe in the previous section, where the density model was only updated \emph{within} an episode. \changes{In this case the model is updated after a collection of episodes}. This makes the objective for RL somewhat non-stationary \changes{and could theoretically cause issues for convergence}, however we found in practice that the increased representational capacity provides significant improvement in performance.

\section{Evaluation Environments}
\label{sec:envs}

\begin{figure*}[htb]
\scriptsize
\centering
\includegraphics[trim={0.0cm 0.0cm 0.0cm 0.0cm},clip,width=0.15\linewidth]{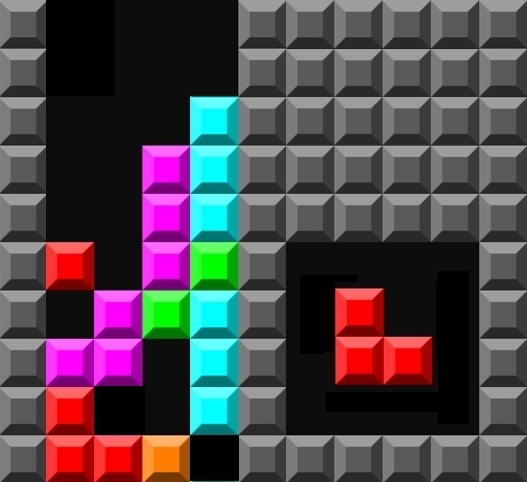}
\includegraphics[trim={0.0cm 0.0cm 0.0cm 0.0cm},clip,width=0.18\linewidth]{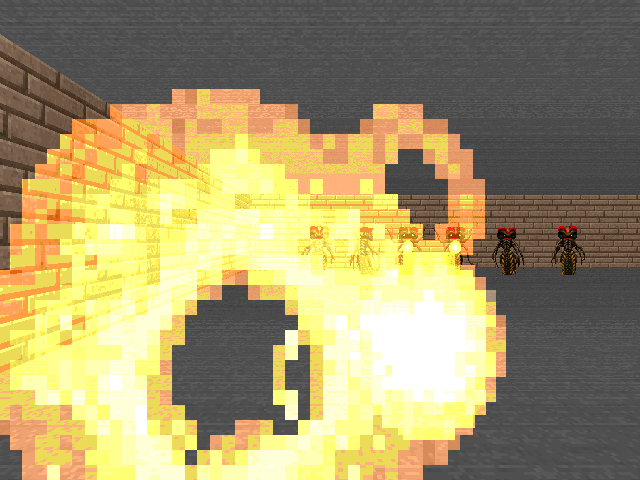}
\includegraphics[trim={0.0cm 0.0cm 0.0cm 0.0cm},clip,width=0.18\linewidth]{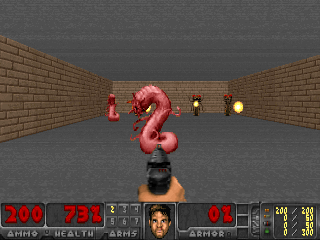}
\includegraphics[trim={0.0cm 0.0cm 0.0cm 0.0cm},clip,width=0.14\linewidth]{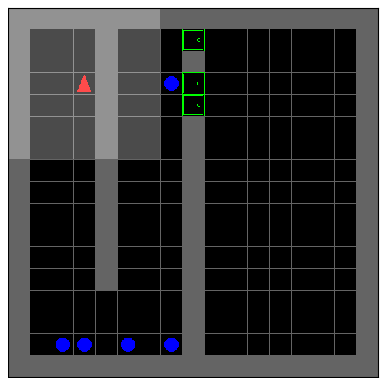}
\\
\includegraphics[trim={8.0cm 1.0cm 0.0cm 4.0cm},clip,width=0.20\linewidth]{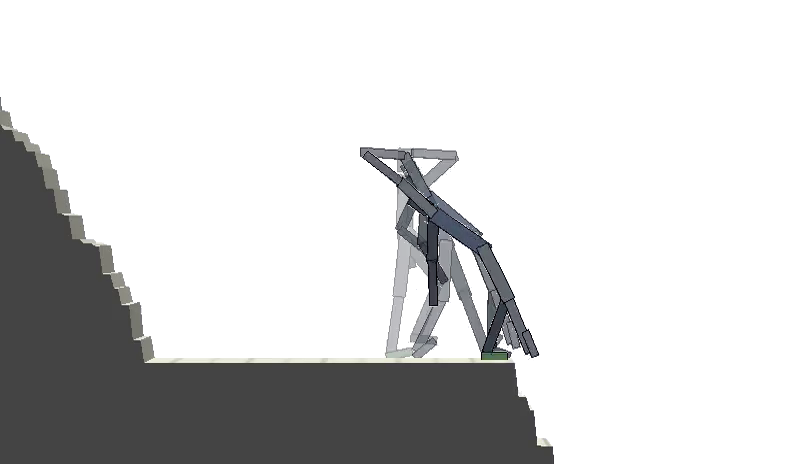}
\includegraphics[trim={8.0cm 0.0cm 0.0cm 5.0cm},clip,width=0.20\linewidth]{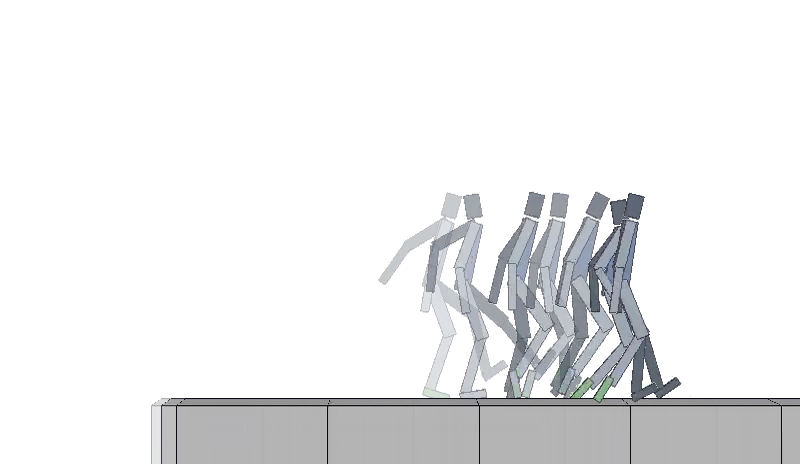}
\includegraphics[trim={15.0cm 4.0cm 21.0cm 10.0cm},clip,width=0.145\linewidth]{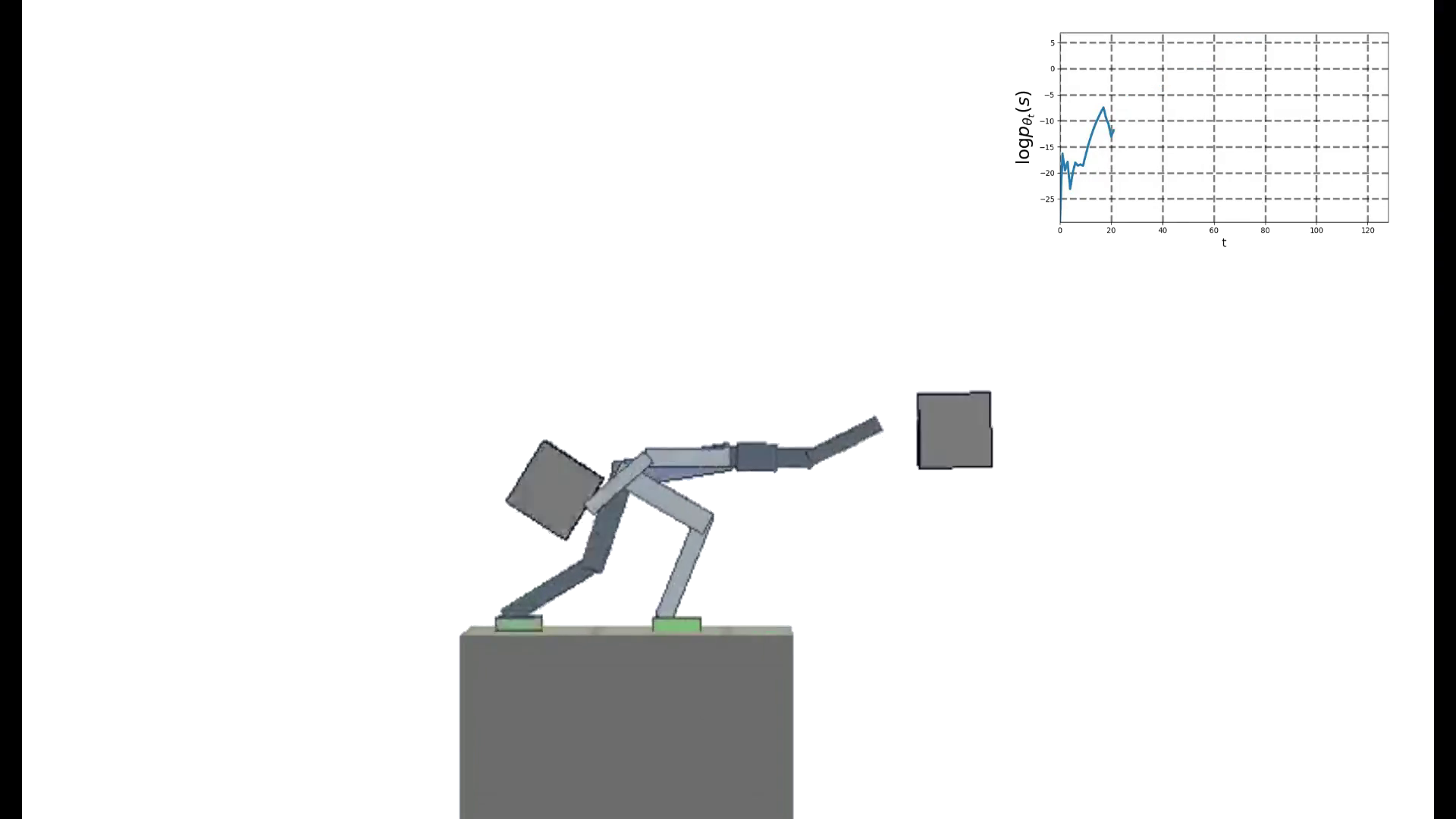}
\includegraphics[trim={8.0cm 0.0cm 0.0cm 5.0cm},clip,width=0.20\linewidth]{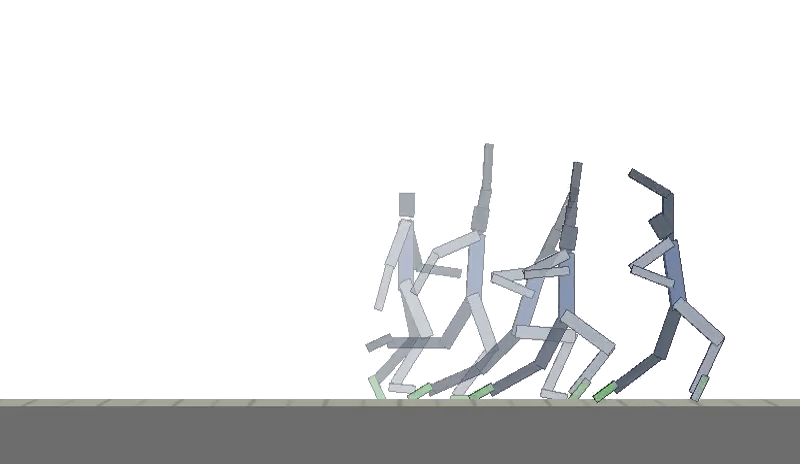}
\caption{
Evaluation environments. Top row, left to right: \Tetris environment, \VizDoom \VizDoomTakeCover and \VizDoomDefendTheLine,
\miniGrid with pursuing ``enemies,'' where the agent can reach a more stable state by finding the doors and leaving the region with enemies. Bottom row, left to right:
\humanoid next to a \humanoidCliff, \humanoid on a \humanoidTreadmill, \humanoidPedestal, \humanoid learning to walk. 
}
\label{fig:discreteenvs}
\vspace{-0.5cm}
\end{figure*}

We evaluate \methodName on a range of environments, from video game domains to simulated robotic control scenarios. In these \entropic environments, the world evolves automatically, without the goal-driven behavior of the agent, due to disruptive forces and adversaries. Standard RL benchmark tasks are typically static, in the sense that unexpected events don not happen unless the agent carries out a specific and coordinated sequence of actions. We therefore selected these environments specifically to be \emph{\entropic}, as we discuss below.
This section describes each environment, with details of the corresponding MDPs in Appendix~\ref{sec:implementation}. Illustrations of the environments are shown in Figure~\ref{fig:discreteenvs}.

\noindent \textbf{\Tetris.}
The classic game offers a naturally \entropic environment --- the world evolves according to its own dynamics even in the absence of coordinated agent actions, piling pieces and filling the board. The agent's task is to place randomly supplied blocks to construct and eliminate complete rows. \changes{The environment gives a reward of $-1$ when the agent fails or dies by stacking a column too high. Otherwise, the agent gets 0.}

\noindent \textbf{\VizDoom.}
We consider two \VizDoom environments from \cite{kempka2016vizdoom}: \VizDoomTakeCover and \VizDoomDefendTheLine where enemies throw fireballs at the agent, which can move around to avoid damage.
\VizDoomTakeCover is \entropic and evolving, with new enemies appearing over time and firing at the player. 
The agent is evaluated on how many fireballs hit it, which we term the ``damage" taken by the agent.

\textbf{\miniGrid.} This is a partially observed navigation task.
The agent (red) starts on the left of the map, and is pursued by ``enemies"
(blue). To escape, the agent can navigate down the hallways and through randomly placed doors (green) to reach the \textit{safe} room on the right, which the enemies cannot enter. To get to the \textit{safe} room the agent must endure increased surprise early on, since the doors appear in different locations in each episode.

\textbf{Simulated \humanoid robots.}
A simulated planar \humanoid agent must avoid falling in the face of external disturbances~\citep{DBLP:journals/corr/abs-1804-06424}. %
We evaluate four versions of this task. %
For \humanoidCliff the agent is initialized sliding towards a cliff, for \humanoidTreadmill, the agent is on a small platform moving backwards at \valueWithUnits{$1$}{m/s}. %
In \humanoidPedestal, random forces are applied to it, and objects are thrown at it.
In \humanoidWalk, we evaluate how the \methodName reward stabilizes an agent that is learning to walk.
In all four tasks, we evaluate the proportion of episodes the robot does not fall. %

\textbf{Training Details.} For discrete action environments, the RL algorithm used is DQN~\citep{mnih2013playing} with a target network. For the \humanoid domains, we use TRPO~\citep{schulman2015trust}. For \Tetris and the \humanoid domains, the policies are parameterized by fully connected neural networks, while \VizDoom uses a convolutional network. Additional details are in Appendix~\refSection{sec:implementation}.

\begin{wraptable}{R}{0.5\textwidth}
	\vspace{-0.5cm}
	\small
	\centering
	\begin{tabular}{lccc}
		\toprule
		Environment   & RND* & SMiRL* & Relative \\ 
		\midrule
		\VizDoomDefendTheLine & -0.3$\pm 0.6$ & -43.1$\pm 0.4$ & -43.4\\
		\textit{Tetris} & 1.5$\pm 2.7$ & -11.9$\pm 2.1$ & -10.4 \\ 
		\VizDoomTakeCover & -1.2$\pm 0.7$ & -7.3$\pm 0.7$ & -8.5 \\ 
		\hline
		\textit{Assault} & 11.3$\pm 1.4$ & -56.9$\pm 2.3$ & -45.6 \\
		\textit{SpaceInvaders} & 1.9$\pm 3.4$ & -10.2$\pm 4.2$ & -8.3 \\
		\textit{Carnival} & 20.4$\pm 1.4$ & -23.1$\pm 4.3$ & -2.7 \\
		\textit{RiverRaid} & -5.5$\pm 3.4$ & 5.8$\pm 3.2$ & 0.3 \\
		\textit{Gravitar} & 30.8$\pm 1.7$ & -26.5$\pm 1.3$ & 4.3 \\ 
		\textit{Berzerk} & 17.2$\pm 1.4$ & -2.9$\pm 4.7$ & 14.3 \\ 
		\bottomrule
	\end{tabular}
	\caption{\small \changes{Difference in entropy vs. a \textit{Random} policy (SMiRL*=SMiRL-Random and RND*=RND-Random, Relative=RND*+SMiRL*).
	More negative values indicate more \entropic environments.} \changes{Note the  \emph{negative} relative entropy gap on our tasks and for \textit{Assault} and \textit{SpaceInvaders}.}}
	\label{tab:entropy-compare-table}
	\vspace{-0.5cm}
\end{wraptable}
\label{sec:dynamic-envs}
\noindent \textbf{Environment Stability.} In \refSection{sec:method}, we described the connection between \methodName and \entropic environments. \changes{We can quantify how \textit{\entropic} an environment is by computing a \textit{relative entropy gap}. We compare the entropy between three methods: entropy minimizing (SMiRL), entropy maximizing (RND) methods, and an initial random (Random) policy (or, more generally, an uninformed policy, such as a randomly initialized neural network).} In stable environments, an uninformed random policy would only attain slightly higher state entropy than one that minimizes the entropy explicitly \changes{(SMiRL - Random $\sim 0$)} , whereas a novelty-seeking policy should attain much higher entropy \changes{(RND - Random $> 0$)}, indicating a relative entropy gap in the \emph{positive} direction. In an \entropic environment, we expect random policies and novelty-seeking policies should attain similar entropies, whereas entropy minimization should result in much lower entropy \changes{(SMiRL - Rand $< 0$)}, indicating a \emph{negative} entropy gap.
\changes{To compute the entropy used in this evaluation, we used the approximation in \refEquation{eq:smirl-ent-approx} multiplied by $-1$} for three of our tasks as well as many Atari games studied in the RND paper~\citep{burda2018rnd}, with numbers shown in \refTable{tab:entropy-compare-table} and full results in Appendix~\ref{sec:full_stability_env_analysis}.
Our environments have a large \emph{negative} entropy gap, whereas most Atari games lack this clear entropy gap.\footnote{We expect that in all cases, random policies will have somewhat higher state entropy than \methodName, so the entropy gap should be interpreted in a relative sense.} We therefore expect \methodName to perform well on these tasks, which we use in the next section, but poorly on most Atari games. We show animations of the resulting policies on our anonymous project \href{https://sites.google.com/view/surpriseminimization}{website}.

\section{Experimental Results}
\label{sec:results}

Our experiments aim to answer the following questions: \textbf{(1)} Can \methodName learn meaningful and complex emergent behaviors without supervision?
\textbf{(2)} Can we improve \methodName by incorporating representation learning via VAEs, as described in~\refSection{sec:method}?
\textbf{(3)} Can \methodName serve as a joint training objective to accelerate the acquisition of reward-guided behavior, and does it outperform prior intrinsic motivation methods in this role? We also illustrate several applications of \methodName, showing that it can accelerate task learning, facilitate exploration, and implement a form of imitation learning.
Video results of learned behaviors are available at \url{https://sites.google.com/view/surpriseminimization}

\subsection{Emergent Behavior with Unsupervised Learning}
\begin{figure*}[tb]
	\centering
    {\includegraphics[trim={0.0cm 0.0cm 0.0cm 0.0cm},clip,valign=t,width=0.30\linewidth]{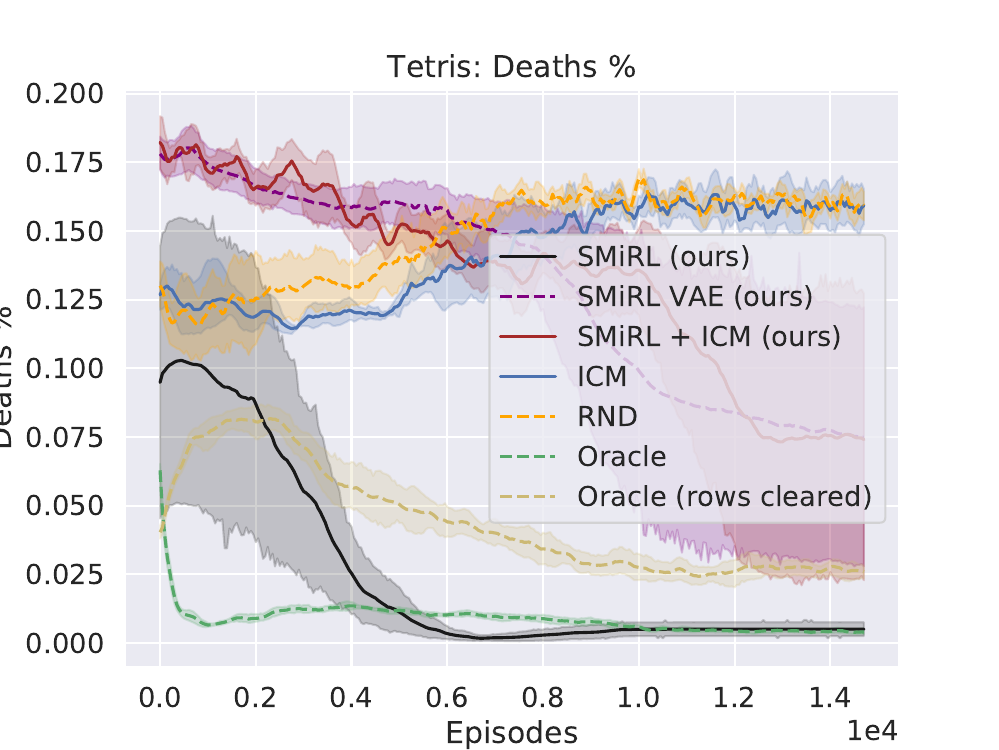}}
    {\includegraphics[trim={0.0cm 0.0cm 0.0cm 0.0cm},clip,valign=t,width=0.30\linewidth]{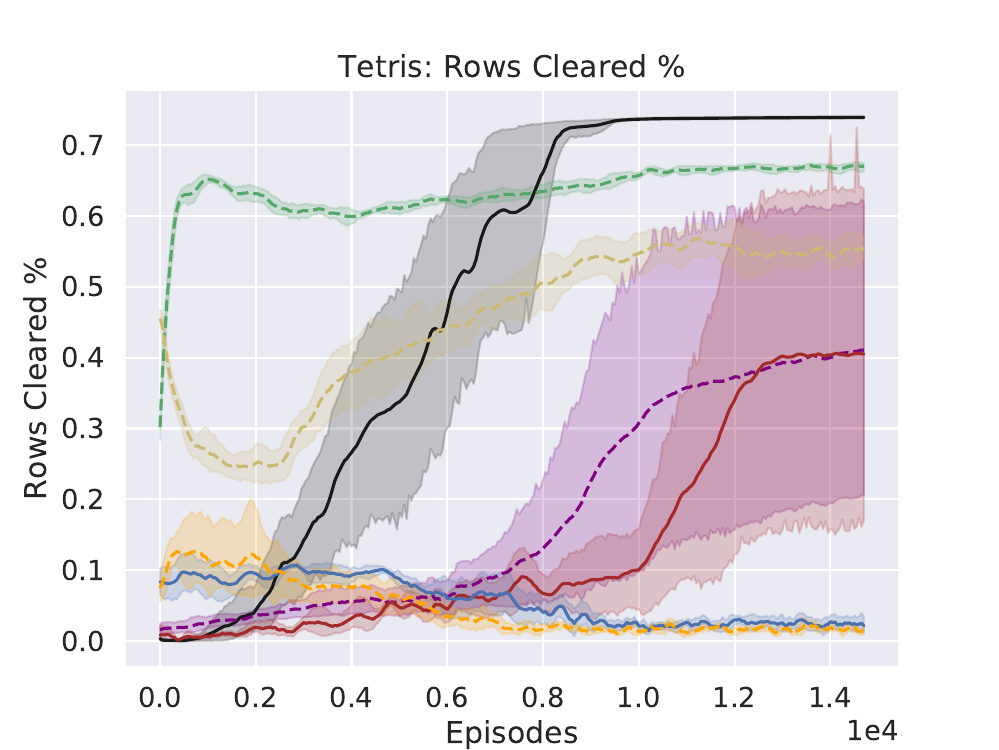}}
    {\includegraphics[trim={0.0cm 0.0cm 0.0cm 0.0cm},clip,valign=t,width=0.28\linewidth]{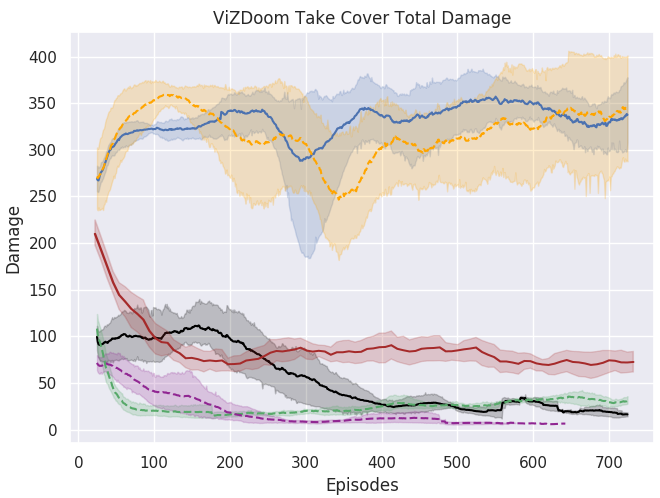}} \\
    
    \includegraphics[trim={0.0cm 0.0cm 0.0cm 0.0cm},clip,width=0.3\linewidth]{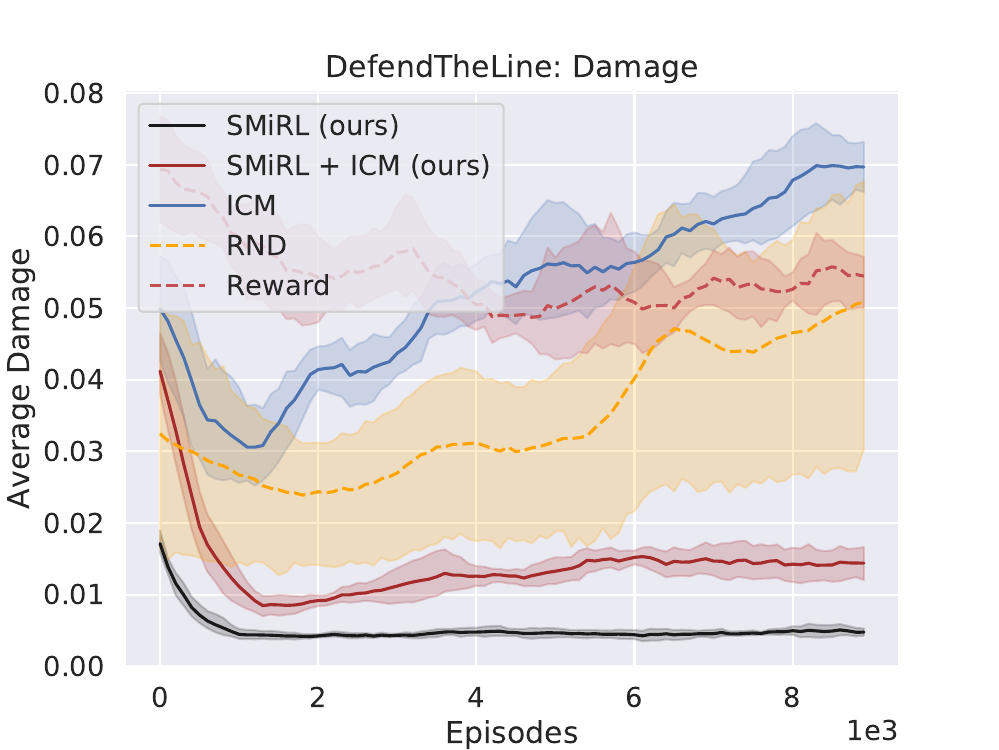}
     \includegraphics[trim={0.0cm 0.0cm 0.0cm 0.0cm},clip,width=0.30\linewidth]{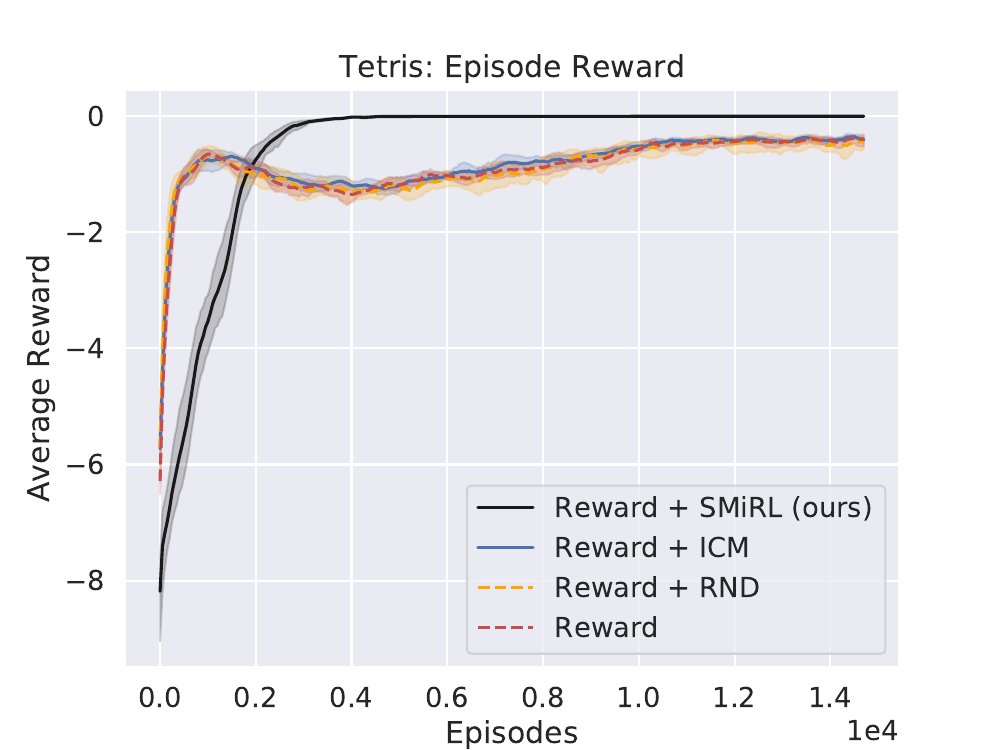}
    \includegraphics[trim={0.0cm 0.0cm 0.0cm 0.0cm},clip,width=0.3\linewidth]{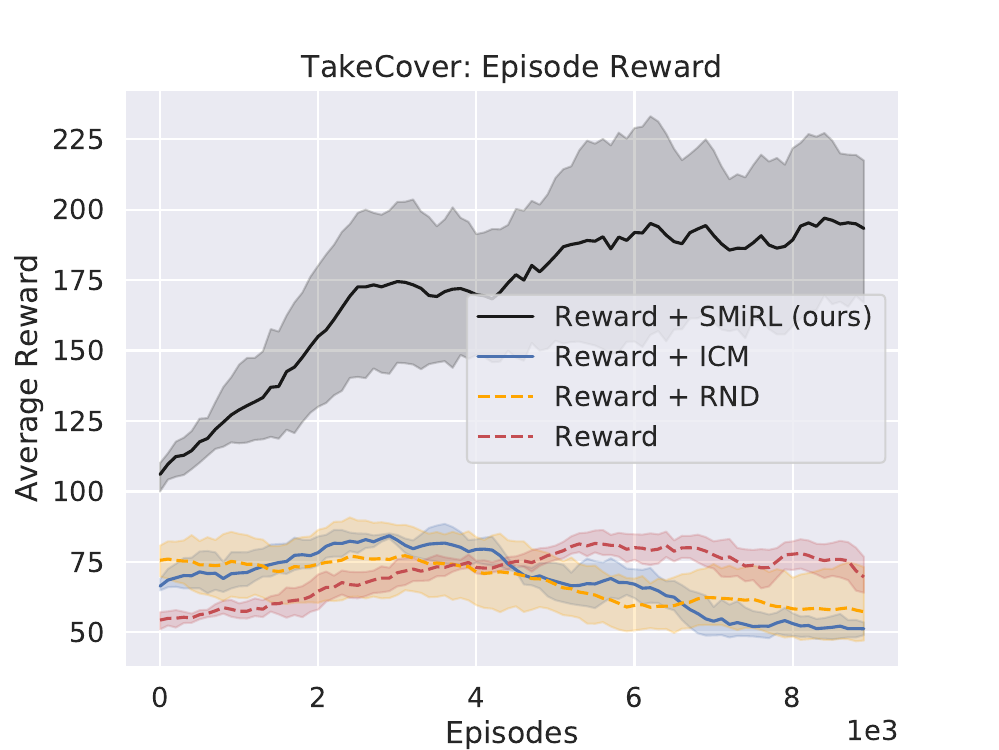}
	\caption{
		 \changes{Comparison between \methodName, ICM, RND, and an Oracle baseline that uses the true reward, evaluated on \Tetris with (top-left) number of deaths per episode (lower is better), (top-center) rows cleared per episode (higher is better), and  in \VizDoomTakeCover (top-right) and \VizDoomDefendTheLine (bottom-left) on amount of damage taken (lower is better). In all cases, the RL algorithm used for training is DQN, and all results are averaged over 6 random seeds,}  \changes{with the shaded areas indicating the standard deviation. In \Tetris (bottom-center) and \VizDoomTakeCover (bottom-right) methods are evaluated on how they improve learning when added to the environment reward function.}
		}\vspace{-0.15in}
	\label{fig:dicrete_results}
\end{figure*}

\begin{figure*}[b]
\centering
\includegraphics[trim={0.0cm 0.0cm 0.0cm 0.0cm},clip,width=0.32\textwidth]{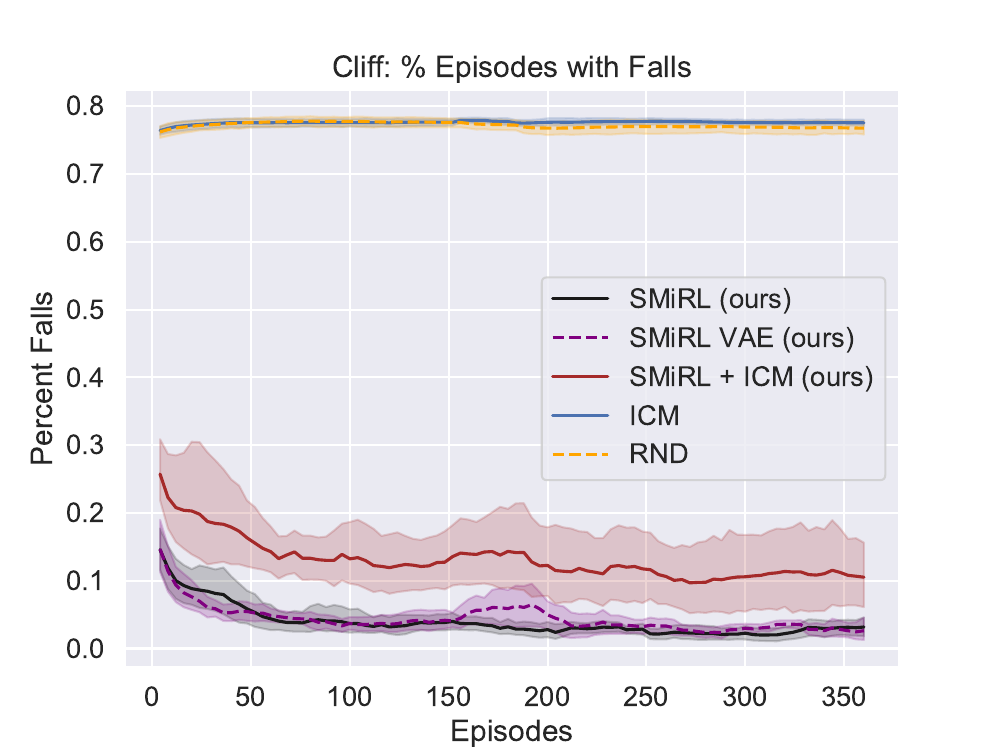}
\includegraphics[trim={0.0cm 0.0cm 0.0cm 0.0cm},clip,width=0.32\textwidth]{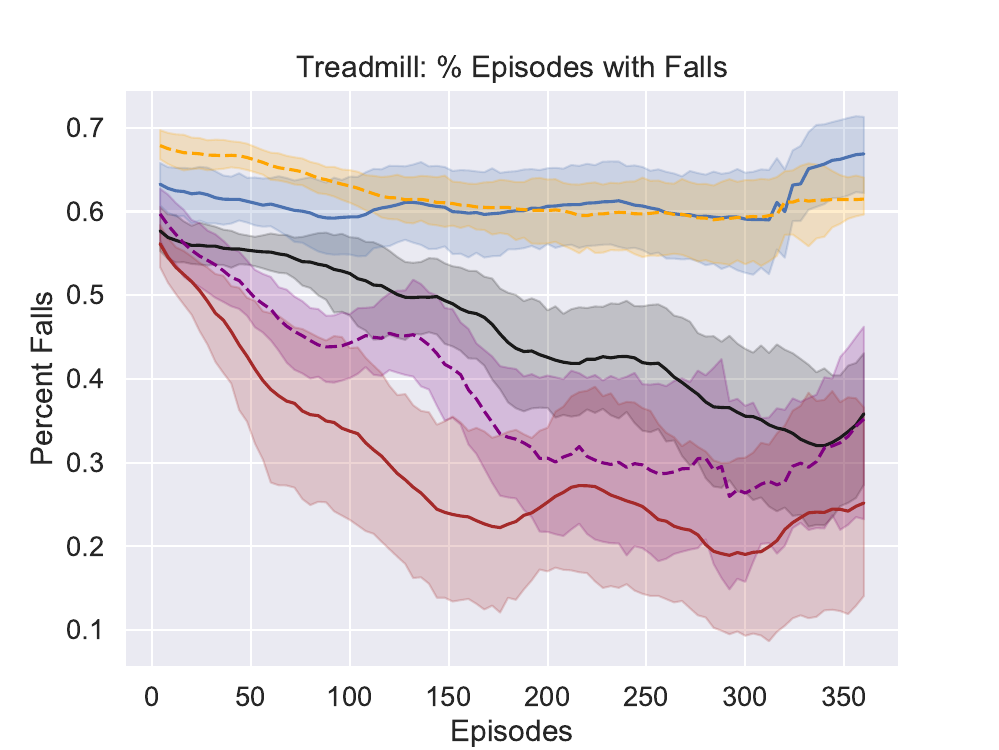}
\includegraphics[trim={0.0cm 0.0cm 0.0cm 0.0cm},clip,width=0.32\linewidth]{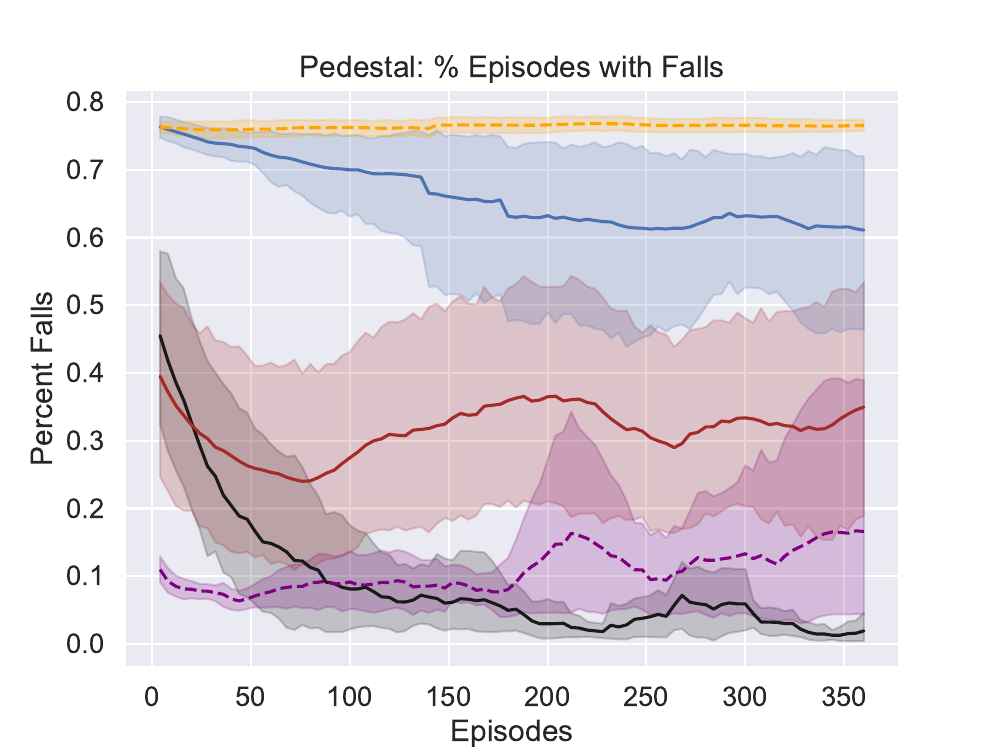}
\caption{\label{fig:biped_curves} \humanoidCliff, \humanoidTreadmill and \humanoidPedestal results. In all cases, \methodName reduces episodes with falls (lower is better). \methodName that uses the VAE for representation learning typically attains better performance. Trained using TRPO with results averaged over 12 random seeds, showing mean and standard deviation \changes{in the shaded area}.
}
\vspace{-0.15in}
\end{figure*}

To answer \textbf{(1)}, we evaluate \methodName on the \Tetris, \VizDoom and \humanoid tasks, studying its ability to generate purposeful coordinated behaviors without engineered task-specific rewards. %
We compare \methodName to two intrinsic motivation methods, ICM~\citep{Pathak2017} and RND~\citep{burda2018rnd}, which seek out states that \emph{maximize} surprise or novelty. %
For reference, we also include an Oracle baseline that directly optimizes the task reward.
We find that \methodName acquires meaningful emergent behaviors across these domains.
In both the \Tetris and \VizDoom environments, stochastic and chaotic events force the \methodName agent to take a coordinated course of action to avoid unusual states, such as full Tetris boards or fireball explosions.
On \Tetris, after training for $3000$ epochs, \methodName achieves near-perfect play, on par with the oracle baseline, with no deaths, \changes{indicating that \methodName may provide better dense rewards than the \textit{Oracle} reward,} as shown in~\refFigure{fig:dicrete_results} (top-left, top-middle). 
\changes{\refFigure{fig:dicrete_results} top-left and top-center show data from the same experiment that plots two different metrics, where the \textit{Oracle} is optimized for minimizing deaths. We include another oracle, \textit{Oracle (rows cleared)} where the reward function is the number of rows cleared.}
ICM and RND seek novelty by creating more and more distinct patterns of blocks rather than clearing them, leading to deteriorating game scores over time. The \methodName agent also learns emergent game playing behavior in \VizDoom, acquiring an effective policy for dodging the fireballs thrown by the enemies, illustrated in~\refFigure{fig:dicrete_results} (top-right and bottom-left). 
Novelty-seeking seeking methods once again yield deteriorating rewards over time. In \humanoidCliff, the \methodName agent learns %
to brace against the ground and stabilize itself at the edge, as shown in~\refFigure{fig:discreteenvs}. In \humanoidTreadmill, \methodName learns to jump forward to increase the time it stays on the treadmill.
In \humanoidPedestal, the agent must actively respond to persistent disturbances. We find that \methodName learns a policy that can reliably keep the agent atop the pedestal, as shown in~\refFigure{fig:discreteenvs}. \refFigure{fig:biped_curves} plots the reduction in falls in the \humanoid environments. %
Novelty-seeking methods learn irregular behaviors that cause the humanoid to jump off the \humanoidCliff and \humanoidPedestal tasks and roll around on the \humanoidTreadmill, maximizing the variety (and quantity) of falls.

Next, we study how representation learning with a VAE improves the \methodName algorithm (question \textbf{(2)}). In these experiments, we train a VAE model and estimate surprise in the VAE latent space. This leads to faster acquisition of the emergent behaviors for \VizDoomTakeCover (\refFigure{fig:dicrete_results}, top-right), \humanoidCliff (\refFigure{fig:biped_curves}, left), and \humanoidTreadmill (\refFigure{fig:biped_curves}, middle), where it also leads to a more successful locomotion behavior.

\begin{figure}[tb]
\centering
\includegraphics[trim={0.0cm 0.0cm 0.0cm 0.0cm},clip,width=0.9\linewidth]{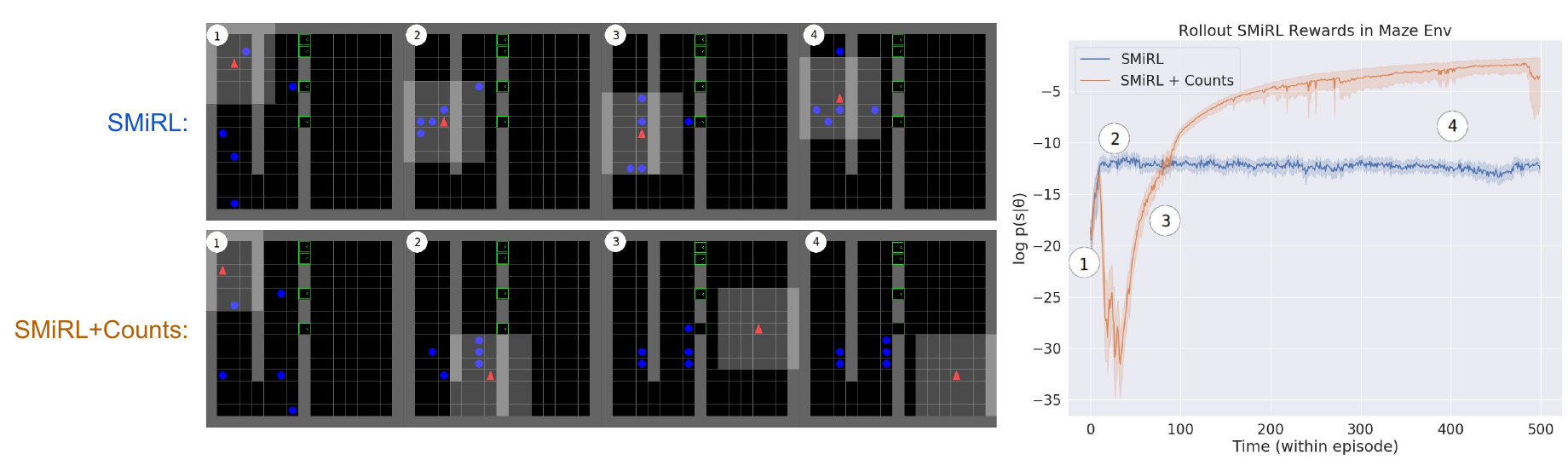}
\caption{\label{fig:miniGrid_curves} 
Here we show \methodName's incentive for longer-term planning \changes{in the \miniGrid environment. On the top-left, we see that \methodName on its own does not explore well enough to reach the \textit{safe} room on the right. Adding exploration via \textit{Counts} (bottom-left) allows \methodName to discover more optimal entropy reducing policies, shown on the right.}
}
\vspace{-0.55cm}
\end{figure}

At first glance, the \methodName surprise minimization objective appears to be the opposite of standard intrinsic motivation objectives~\citep{bellemare2016unifying,Pathak2017,burda2018rnd} that seek out states with \emph{maximal} surprise (i.e., novel states). However, while those approaches measure surprise with respect to all prior experience, \methodName minimizes surprise over each episode. 
We demonstrate that these two approaches are in fact complementary. %
\methodName can use conventional intrinsic motivation methods to aid in exploration \emph{so as to discover more effective policies for minimizing surprise}. We can, therefore, combine these two methods and learn more sophisticated behaviors. %
While \methodName on its own does not successfully produce a good walking gait on \humanoidTreadmill, the addition of novelty-seeking intrinsic motivation allows increased exploration, which results in an improved walking gait that remains on the treadmill longer, as shown in~\refFigure{fig:biped_curves} (middle).
We evaluate this combined approach across environments including \humanoidPedestal and \humanoidCliff as well, where learning to avoid falls is also a challenge. For these two tasks \methodName can already discover strong surprise minimizing policies and adding exploration bonuses does not provide additional benefit.
In~\refFigure{fig:miniGrid_curves} adding a bonus enables the agent to discover improved surprise minimizing strategies.

\textbf{\methodName and long term surprise.} Although the \methodName objective by itself does not specifically encourage exploration, we observe that optimal \methodName policies exhibit active ``searching'' behaviors, seeking out objects in the environment that would allow for reduced long-term surprise. For example, in \miniGrid, the positions of the doors leading to the safe room change between episodes, and the policy trained with \methodName learns to search for the doors to facilitate lower future surprise, even if finding the doors themselves yields higher short-term surprise. This behavior is illustrated in~\refFigure{fig:miniGrid_curves}, along with the ``delayed gratification'' plot, which shows that the \methodName agent incurs higher surprise early in the episode, for the sake of much lower surprise later.

\begin{wrapfigure}{r}{0.3\textwidth}
\vspace{-0.5cm}
\centering
{\scriptsize Targets~|~States attained by \methodName}
\includegraphics[trim={0.0cm 0.0cm 0.0cm 0.0cm},clip,width=0.8\linewidth]{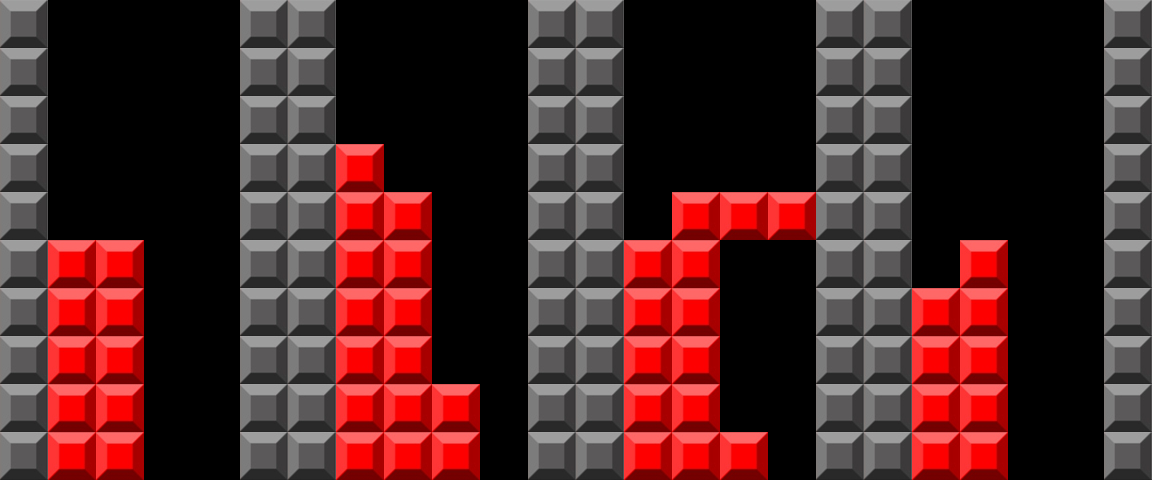} \\
\includegraphics[trim={0.0cm 0.0cm 0.0cm 0.0cm},clip,width=0.8\linewidth]{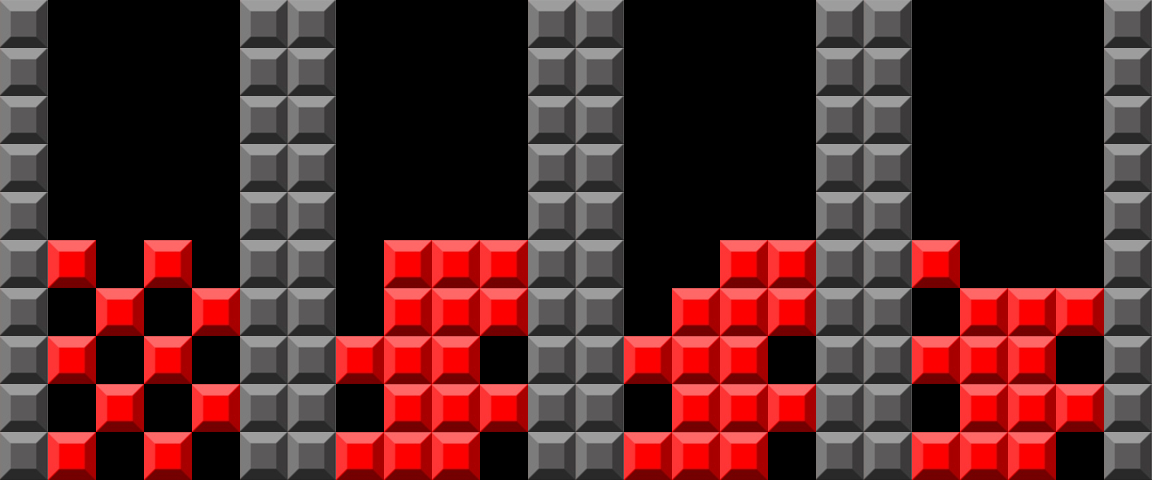}
\caption{ Tetris \textit{imitation} by starting $p_{\theta}(\bs)$ with left image. %
}\label{fig:imitation_rectangle} 
\vspace{-0.05in}
\end{wrapfigure} 
\subsection{Applications of \methodName}
\label{sec:method:IL}
While the focus of this paper is on the emergent behaviors obtained by \methodName, here we study more pragmatic applications. We show that \methodName can be used for basic imitation and joint training to accelerate reward-driven learning.

\textbf{Imitation.}
\methodName can be adapted to perform imitation by initializing the prior via the buffer $\mathcal{D}_0$ with states from demonstrations, or individual desired outcome states. We initialize the buffer $\mathcal{D}_0$ in \Tetris with user-specified \emph{desired} board states. An illustration of the \Tetris imitation task is presented in~\refFigure{fig:imitation_rectangle}, showing imitation of a box pattern (top) and a checkerboard pattern (bottom), with the leftmost frame showing the user-specified example, and the other frames showing actual states reached by the \methodName agent. While several prior works have studied imitation without example actions~\citep{liu2018imitation,torabi2018behavioral,aytar2018playing,torabi2018generative,edwards2018imitating,leestate}, this capability emerges automatically in \methodName, without any further modification to the algorithm.

\begin{figure}%
\vspace{-0.0cm}
\centering
  \includegraphics[trim={0.0cm 0.0cm 0.0cm 0.0cm},clip,width=0.32\linewidth]{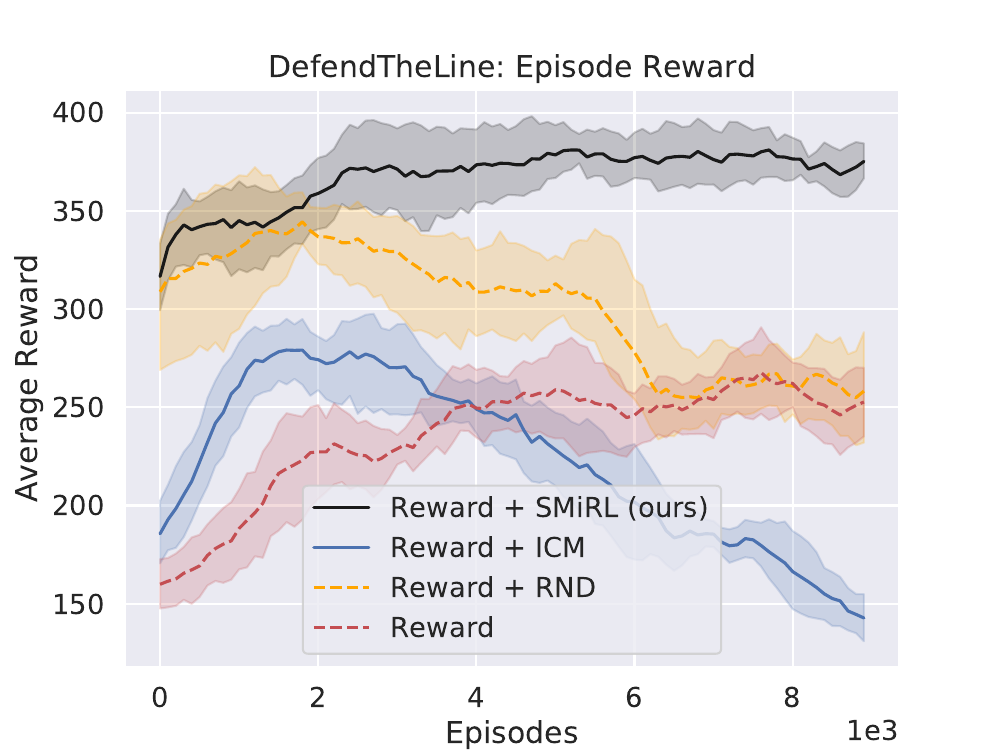}
\includegraphics[trim={1.0cm 0.0cm 1.0cm 0.0cm},clip,width=0.3\linewidth]{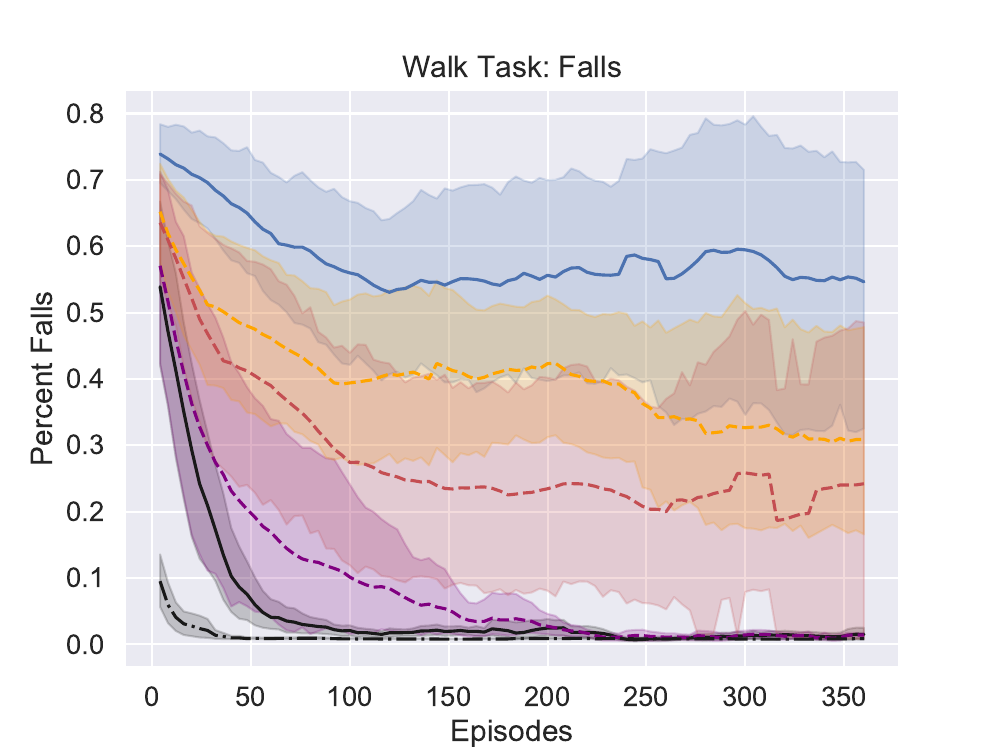}
\includegraphics[trim={1.0cm 0.0cm 1.0cm 0.0cm},clip,width=0.3\linewidth]{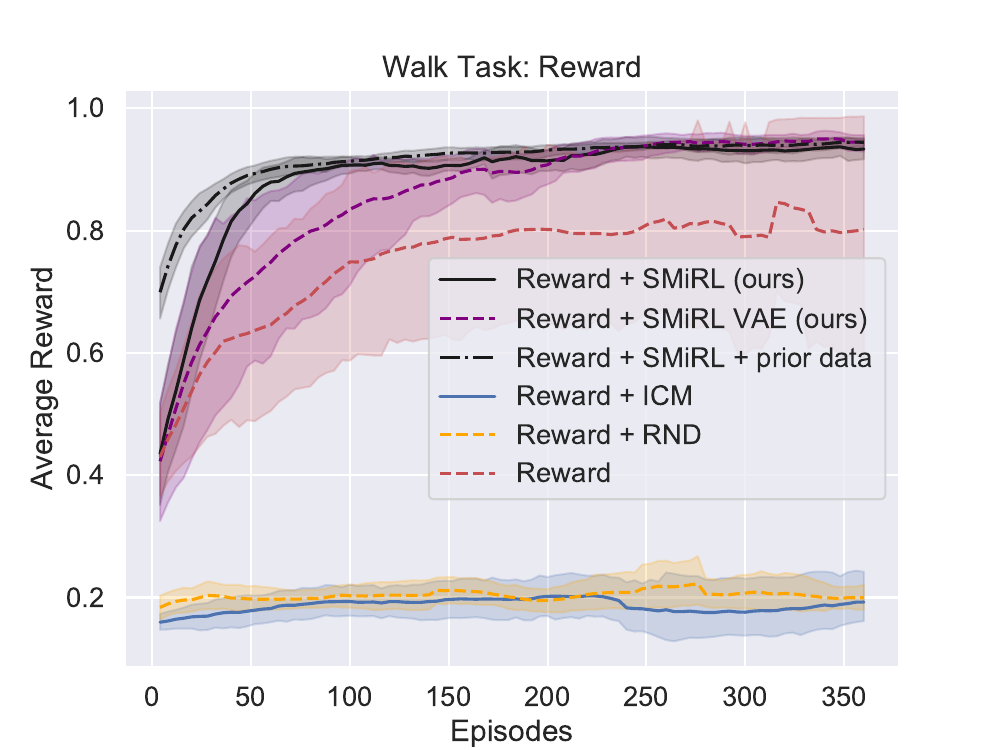}
\caption{ Left: We combine \methodName with the survival time task reward in the \VizDoomDefendTheLine task. Middle/Right:
 We combine the \methodName reward with the \humanoidWalk reward and initialize \methodName without walking prior walking data (ours) and with (prior data). Results over 12 seeds \changes{with standard deviation indicated by the shaded area}.
}
\label{fig:joint_results_bip}
\vspace{-0.5cm}
\end{figure}

\textbf{\methodName as an auxiliary reward.}
We explore how combining \methodName with a task reward can lead to faster learning. We hypothesize that, when the task reward is aligned with avoiding unpredictable situations (e.g., falling or dying), adding \methodName as an auxiliary reward can accelerate learning by providing a dense intermediate signal. The full reward is given by $r_\text{combined}(\bs) = r_\text{task}(\bs) + \alpha r_\text{\methodName}(\bs)$, where $\alpha$ is chosen to put the two reward terms at a similar magnitude. We study this application of \methodName in \changes{the tasks: \Tetris in~\refFigure{fig:dicrete_results}~(bottom-center), \VizDoomTakeCover in~\refFigure{fig:dicrete_results}~(bottom-right), \VizDoomDefendTheLine and \humanoidWalk. On the easier tasks \Tetris and \VizDoomTakeCover task~(\refFigure{fig:joint_results_bip}), prior exploration methods generally lead to significantly worse performance and \methodName improves learning speed. On the harder \humanoidWalk and \VizDoomDefendTheLine tasks, the \methodName reward accelerates learning substantially, and also significantly reduces the number of falls or deaths. We found that increasing the difficulty of \VizDoomTakeCover and \VizDoomDefendTheLine (via the environment's difficulty setting~\citep{kempka2016vizdoom}) resulted in a clearer separation between \methodName and other methods}

In \humanoidWalk, we include a version of \methodName with \emph{prior data}, where $p_\theta(\bs)$ is initialized with $8$ walking trajectories ($256$ timesteps each), similar to the imitation setting. Incorporating prior data requires no modification to the \methodName algorithm, and we can see in \refFigure{fig:joint_results_bip}~(middle and right) that this variant (``Reward + \methodName + prior data'') further accelerates learning and reduces the number of falls. \changes{This shows that while \methodName can learn from scratch, it is possible to encode prior knowledge in $p_\theta(\bs)$ to improve learning.}

\section{Discussion}
\label{sec:discussion}

We presented an unsupervised reinforcement learning method based on \emph{minimizing} surprise. We show that surprise minimization can be used to learn a variety of behaviors that reach ``homeostasis,'' 
putting the agent into stable state distributions in its environment. Across a range of tasks, these cycles correspond to useful, semantically meaningful, and complex behaviors: clearing rows in \Tetris, avoiding fireballs in \VizDoom, and learning to balance and hop with a bipedal robot. The key insight utilized by our method is that, in contrast to simple simulated domains, realistic environments exhibit \entropic phenomena that gradually increase entropy over time. An agent that resists this growth in entropy must take effective and coordinated actions, thus learning increasingly complex behaviors. This stands in contrast to commonly proposed intrinsic exploration methods based on novelty.

Besides fully unsupervised reinforcement learning, where we show that our method can give rise to intelligent and sophisticated policies, we also illustrate several more practical applications of our approach. We show that surprise minimization can provide a general-purpose auxiliary reward
that, when combined with task rewards, can improve learning in environments where avoiding catastrophic (and surprising) outcomes is desirable. We also show that \methodName can be adapted to perform a rudimentary form of imitation.

Our investigation of surprise minimization suggests several directions for future work. The particular behavior of a surprise minimizing agent is strongly influenced by the choice of state representation: by including or excluding particular observation modalities, the agent will be more or less surprised. Thus, tasks may be designed by choosing an appropriate state or observation representations. 
Exploring this direction may lead to new ways of specifying behaviors for RL agents without explicit reward design.
Other applications of surprise minimization may also be explored in future work, possibly for mitigating reward misspecification by disincentivizing any unusual behavior that likely deviates from what the reward designer intended.
\changes{The experiments in this work make use of available or easy to learn state representations. Using these learned representations does not address the difficulty of estimating and minimizing surprise across episodes or more generally over long sequences (possibly a single episode) which is a challenge for surprise minimization-based methods.
We believe that non-episodic surprise minimization is a promising direction for future research to study how  surprise minimization can result in intelligent and sophisticated behavior that maintains homeostasis by acquiring increasingly complex behaviors.}

\bibliographystyle{iclr2021_conference}
\bibliography{references} 

\begin{thebibliography}{53}
\providecommand{\natexlab}[1]{#1}
\providecommand{\url}[1]{\texttt{#1}}
\expandafter\ifx\csname urlstyle\endcsname\relax
  \providecommand{\doi}[1]{doi: #1}\else
  \providecommand{\doi}{doi: \begingroup \urlstyle{rm}\Url}\fi

\bibitem[Achiam \& Sastry(2017{\natexlab{a}})Achiam and
  Sastry]{DBLP:journals/corr/AchiamS17}
Joshua Achiam and Shankar Sastry.
\newblock Surprise-based intrinsic motivation for deep reinforcement learning.
\newblock \emph{CoRR}, abs/1703.01732, 2017{\natexlab{a}}.
\newblock URL \url{http://arxiv.org/abs/1703.01732}.

\bibitem[Achiam \& Sastry(2017{\natexlab{b}})Achiam and
  Sastry]{achiam2017surprise}
Joshua Achiam and Shankar Sastry.
\newblock Surprise-based intrinsic motivation for deep reinforcement learning.
\newblock \emph{arXiv preprint arXiv:1703.01732}, 2017{\natexlab{b}}.

\bibitem[Annabi et~al.(2020)Annabi, Pitti, and Quoy]{annabi2020autonomous}
Louis Annabi, Alexandre Pitti, and Mathias Quoy.
\newblock Autonomous learning and chaining of motor primitives using the free
  energy principle.
\newblock \emph{arXiv preprint arXiv:2005.05151}, 2020.

\bibitem[Aytar et~al.(2018)Aytar, Pfaff, Budden, Paine, Wang, and
  de~Freitas]{aytar2018playing}
Yusuf Aytar, Tobias Pfaff, David Budden, Thomas Paine, Ziyu Wang, and Nando
  de~Freitas.
\newblock Playing hard exploration games by watching youtube.
\newblock In \emph{Advances in Neural Information Processing Systems}, pp.\
  2930--2941, 2018.

\bibitem[Baker et~al.(2019)Baker, Kanitscheider, Markov, Wu, Powell, McGrew,
  and Mordatch]{baker2019emergent}
Bowen Baker, Ingmar Kanitscheider, Todor Markov, Yi~Wu, Glenn Powell, Bob
  McGrew, and Igor Mordatch.
\newblock Emergent tool use from multi-agent autocurricula.
\newblock \emph{arXiv preprint arXiv:1909.07528}, 2019.

\bibitem[Bansal et~al.(2017)Bansal, Pachocki, Sidor, Sutskever, and
  Mordatch]{bansal2017emergent}
Trapit Bansal, Jakub Pachocki, Szymon Sidor, Ilya Sutskever, and Igor Mordatch.
\newblock Emergent complexity via multi-agent competition.
\newblock \emph{arXiv preprint arXiv:1710.03748}, 2017.

\bibitem[Bellemare et~al.(2016)Bellemare, Srinivasan, Ostrovski, Schaul,
  Saxton, and Munos]{bellemare2016unifying}
Marc Bellemare, Sriram Srinivasan, Georg Ostrovski, Tom Schaul, David Saxton,
  and Remi Munos.
\newblock Unifying count-based exploration and intrinsic motivation.
\newblock In \emph{Advances in Neural Information Processing Systems}, pp.\
  1471--1479, 2016.

\bibitem[Berseth et~al.(2018)Berseth, Peng, and van~de
  Panne]{DBLP:journals/corr/abs-1804-06424}
Glen Berseth, Xue~Bin Peng, and Michiel van~de Panne.
\newblock Terrain {RL} simulator.
\newblock \emph{CoRR}, abs/1804.06424, 2018.
\newblock URL \url{http://arxiv.org/abs/1804.06424}.

\bibitem[Biehl et~al.(2018)Biehl, Guckelsberger, Salge, Smith, and
  Polani]{Biehl2018FreeE}
M.~Biehl, C.~Guckelsberger, C.~Salge, S.~Smith, and D.~Polani.
\newblock Free energy , empowerment , and predictive information compared.
\newblock 2018.

\bibitem[Boltzmann(1886)]{boltzmann1886second}
Ludwig Boltzmann.
\newblock The second law of thermodynamics.
\newblock 1886.

\bibitem[Burda et~al.(2018{\natexlab{a}})Burda, Edwards, Pathak, Storkey,
  Darrell, and Efros]{Burda2018}
Yuri Burda, Harri Edwards, Deepak Pathak, Amos Storkey, Trevor Darrell, and
  Alexei~A. Efros.
\newblock {Large-Scale Study of Curiosity-Driven Learning}.
\newblock 2018{\natexlab{a}}.
\newblock URL \url{http://arxiv.org/abs/1808.04355}.

\bibitem[Burda et~al.(2018{\natexlab{b}})Burda, Edwards, Storkey, and
  Klimov]{burda2018rnd}
Yuri Burda, Harrison Edwards, Amos Storkey, and Oleg Klimov.
\newblock Exploration by random network distillation.
\newblock \emph{ICLR}, 2018{\natexlab{b}}.

\bibitem[Chen et~al.(2020)Chen, Song, Lipson, and Vondrick]{chen2020visual}
Boyuan Chen, Shuran Song, Hod Lipson, and Carl Vondrick.
\newblock Visual hide and seek.
\newblock In \emph{Artificial Life Conference Proceedings}, pp.\  645--655. MIT
  Press, 2020.

\bibitem[Chentanez et~al.(2005)Chentanez, Barto, and
  Singh]{chentanez2005intrinsically}
Nuttapong Chentanez, Andrew~G Barto, and Satinder~P Singh.
\newblock Intrinsically motivated reinforcement learning.
\newblock In \emph{Advances in neural information processing systems}, pp.\
  1281--1288, 2005.

\bibitem[Chevalier-Boisvert et~al.(2018)Chevalier-Boisvert, Willems, and
  Pal]{gym_minigrid}
Maxime Chevalier-Boisvert, Lucas Willems, and Suman Pal.
\newblock Minimalistic gridworld environment for openai gym.
\newblock \url{https://github.com/maximecb/gym-minigrid}, 2018.

\bibitem[Edwards et~al.(2018)Edwards, Sahni, Schroecker, and
  Isbell]{edwards2018imitating}
Ashley~D Edwards, Himanshu Sahni, Yannick Schroecker, and Charles~L Isbell.
\newblock Imitating latent policies from observation.
\newblock \emph{arXiv preprint arXiv:1805.07914}, 2018.

\bibitem[Faraji et~al.(2018)Faraji, Preuschoff, and
  Gerstner]{faraji2018balancing}
Mohammadjavad Faraji, Kerstin Preuschoff, and Wulfram Gerstner.
\newblock Balancing new against old information: the role of puzzlement
  surprise in learning.
\newblock \emph{Neural computation}, 30\penalty0 (1):\penalty0 34--83, 2018.

\bibitem[Friston(2009)]{friston2009free}
Karl Friston.
\newblock The free-energy principle: a rough guide to the brain?
\newblock \emph{Trends in cognitive sciences}, 13\penalty0 (7):\penalty0
  293--301, 2009.

\bibitem[Friston et~al.(2016)Friston, FitzGerald, Rigoli, Schwartenbeck,
  Pezzulo, et~al.]{friston2016active}
Karl Friston, Thomas FitzGerald, Francesco Rigoli, Philipp Schwartenbeck,
  Giovanni Pezzulo, et~al.
\newblock Active inference and learning.
\newblock \emph{Neuroscience \& Biobehavioral Reviews}, 68:\penalty0 862--879,
  2016.

\bibitem[Friston et~al.(2009)Friston, Daunizeau, and
  Kiebel]{10.1371/journal.pone.0006421}
Karl~J. Friston, Jean Daunizeau, and Stefan~J. Kiebel.
\newblock Reinforcement learning or active inference?
\newblock \emph{PLOS ONE}, 4\penalty0 (7):\penalty0 1--13, 07 2009.
\newblock \doi{10.1371/journal.pone.0006421}.
\newblock URL \url{https://doi.org/10.1371/journal.pone.0006421}.

\bibitem[Hazan et~al.(2019)Hazan, Kakade, Singh, and
  Van~Soest]{hazan2019provably}
Elad Hazan, Sham Kakade, Karan Singh, and Abby Van~Soest.
\newblock Provably efficient maximum entropy exploration.
\newblock In \emph{International Conference on Machine Learning}, pp.\
  2681--2691, 2019.

\bibitem[Houthooft et~al.(2016)Houthooft, Chen, Duan, Schulman, {De Turck}, and
  Abbeel]{Houthooft2016}
Rein Houthooft, Xi~Chen, Yan Duan, John Schulman, Filip {De Turck}, and Pieter
  Abbeel.
\newblock {VIME: Variational Information Maximizing Exploration}.
\newblock 2016.
\newblock URL \url{http://arxiv.org/abs/1605.09674}.

\bibitem[Kempka et~al.(2016)Kempka, Wydmuch, Runc, Toczek, and
  Ja{\'s}kowski]{kempka2016vizdoom}
Micha{\l} Kempka, Marek Wydmuch, Grzegorz Runc, Jakub Toczek, and Wojciech
  Ja{\'s}kowski.
\newblock Vizdoom: A doom-based ai research platform for visual reinforcement
  learning.
\newblock In \emph{2016 IEEE Conference on Computational Intelligence and Games
  (CIG)}, pp.\  1--8. IEEE, 2016.

\bibitem[Kim et~al.(2020)Kim, Sano, De~Freitas, Haber, and
  Yamins]{kim2020active}
Kuno Kim, Megumi Sano, Julian De~Freitas, Nick Haber, and Daniel Yamins.
\newblock Active world model learning with progress curiosity.
\newblock \emph{arXiv preprint arXiv:2007.07853}, 2020.

\bibitem[Kim et~al.(2019)Kim, Nam, Kim, Kim, and Kim]{kim2019curiosity}
Youngjin Kim, Wontae Nam, Hyunwoo Kim, Ji-Hoon Kim, and Gunhee Kim.
\newblock Curiosity-bottleneck: Exploration by distilling task-specific
  novelty.
\newblock In \emph{International Conference on Machine Learning}, pp.\
  3379--3388, 2019.

\bibitem[Kingma \& Welling(2014)Kingma and Welling]{kingma2013auto}
Diederik~P Kingma and Max Welling.
\newblock Auto-encoding variational bayes.
\newblock \emph{ICLR}, 2014.

\bibitem[Klyubin et~al.(2005)Klyubin, Polani, and Nehaniv]{10.1007/11553090_75}
Alexander~S. Klyubin, Daniel Polani, and Chrystopher~L. Nehaniv.
\newblock All else being equal be empowered.
\newblock In Mathieu~S. Capcarr{\`e}re, Alex~A. Freitas, Peter~J. Bentley,
  Colin~G. Johnson, and Jon Timmis (eds.), \emph{Advances in Artificial Life},
  pp.\  744--753, Berlin, Heidelberg, 2005. Springer Berlin Heidelberg.
\newblock ISBN 978-3-540-31816-3.

\bibitem[Lee et~al.()Lee, Eysenbach, Parisotto, Salakhutdinov, and
  Levine]{leestate}
Lisa Lee, Benjamin Eysenbach, Emilio Parisotto, Ruslan Salakhutdinov, and
  Sergey Levine.
\newblock State marginal matching with mixtures of policies.

\bibitem[Lee et~al.(2019)Lee, Eysenbach, Parisotto, Xing, Levine, and
  Salakhutdinov]{lee2019efficient}
Lisa Lee, Benjamin Eysenbach, Emilio Parisotto, Eric Xing, Sergey Levine, and
  Ruslan Salakhutdinov.
\newblock Efficient exploration via state marginal matching.
\newblock \emph{arXiv preprint arXiv:1906.05274}, 2019.

\bibitem[Lehman \& Stanley(2011)Lehman and Stanley]{lehman2011abandoning}
Joel Lehman and Kenneth~O Stanley.
\newblock Abandoning objectives: Evolution through the search for novelty
  alone.
\newblock \emph{Evolutionary computation}, 19\penalty0 (2):\penalty0 189--223,
  2011.

\bibitem[Liu et~al.(2018)Liu, Gupta, Abbeel, and Levine]{liu2018imitation}
YuXuan Liu, Abhishek Gupta, Pieter Abbeel, and Sergey Levine.
\newblock Imitation from observation: Learning to imitate behaviors from raw
  video via context translation.
\newblock In \emph{2018 IEEE International Conference on Robotics and
  Automation (ICRA)}, pp.\  1118--1125. IEEE, 2018.

\bibitem[Lopes et~al.(2012)Lopes, Lang, Toussaint, and
  Oudeyer]{lopes2012exploration}
Manuel Lopes, Tobias Lang, Marc Toussaint, and Pierre-Yves Oudeyer.
\newblock Exploration in model-based reinforcement learning by empirically
  estimating learning progress.
\newblock In \emph{Advances in neural information processing systems}, pp.\
  206--214, 2012.

\bibitem[Mnih et~al.(2013)Mnih, Kavukcuoglu, Silver, Graves, Antonoglou,
  Wierstra, and Riedmiller]{mnih2013playing}
Volodymyr Mnih, Koray Kavukcuoglu, David Silver, Alex Graves, Ioannis
  Antonoglou, Daan Wierstra, and Martin Riedmiller.
\newblock Playing atari with deep reinforcement learning.
\newblock \emph{arXiv preprint arXiv:1312.5602}, 2013.

\bibitem[Mohamed \& Jimenez~Rezende(2015)Mohamed and
  Jimenez~Rezende]{NIPS2015_5668}
Shakir Mohamed and Danilo Jimenez~Rezende.
\newblock Variational information maximisation for intrinsically motivated
  reinforcement learning.
\newblock In C.~Cortes, N.~D. Lawrence, D.~D. Lee, M.~Sugiyama, and R.~Garnett
  (eds.), \emph{Advances in Neural Information Processing Systems 28}, pp.\
  2125--2133. Curran Associates, Inc., 2015.

\bibitem[Oudeyer \& Kaplan(2009)Oudeyer and Kaplan]{oudeyer2009intrinsic}
Pierre-Yves Oudeyer and Frederic Kaplan.
\newblock What is intrinsic motivation? a typology of computational approaches.
\newblock \emph{Frontiers in neurorobotics}, 1:\penalty0 6, 2009.

\bibitem[Oudeyer et~al.(2007)Oudeyer, Kaplan, and Hafner]{oudeyer2007intrinsic}
Pierre-Yves Oudeyer, Frdric Kaplan, and Verena~V Hafner.
\newblock Intrinsic motivation systems for autonomous mental development.
\newblock \emph{IEEE transactions on evolutionary computation}, 11\penalty0
  (2):\penalty0 265--286, 2007.

\bibitem[Pathak et~al.(2017)Pathak, Agrawal, Efros, and Darrell]{Pathak2017}
Deepak Pathak, Pulkit Agrawal, Alexei~A Efros, and Trevor Darrell.
\newblock {Curiosity-driven Exploration by Self-supervised Prediction}.
\newblock 2017.

\bibitem[Pathak et~al.(2019)Pathak, Gandhi, and Gupta]{Pathak2019}
Deepak Pathak, Dhiraj Gandhi, and Abhinav Gupta.
\newblock {Self-Supervised Exploration via Disagreement}.
\newblock 2019.

\bibitem[Schmidhuber(1991)]{schmidhuber1991curious}
J{\"u}rgen Schmidhuber.
\newblock Curious model-building control systems.
\newblock In \emph{Proc. international joint conference on neural networks},
  pp.\  1458--1463, 1991.

\bibitem[Schneider \& Kay(1994)Schneider and Kay]{schneider1994life}
Eric~D Schneider and James~J Kay.
\newblock Life as a manifestation of the second law of thermodynamics.
\newblock \emph{Mathematical and computer modelling}, 19\penalty0
  (6-8):\penalty0 25--48, 1994.

\bibitem[Schr{\"o}dinger(1944)]{schrodinger1944life}
Erwin Schr{\"o}dinger.
\newblock \emph{What is life? The physical aspect of the living cell and mind}.
\newblock Cambridge University Press Cambridge, 1944.

\bibitem[Schulman et~al.(2015)Schulman, Levine, Abbeel, Jordan, and
  Moritz]{schulman2015trust}
John Schulman, Sergey Levine, Pieter Abbeel, Michael Jordan, and Philipp
  Moritz.
\newblock Trust region policy optimization.
\newblock In \emph{International conference on machine learning}, pp.\
  1889--1897, 2015.

\bibitem[Shyam et~al.(2018)Shyam, Ja{\'s}kowski, and Gomez]{shyam2018model}
Pranav Shyam, Wojciech Ja{\'s}kowski, and Faustino Gomez.
\newblock Model-based active exploration.
\newblock \emph{arXiv preprint arXiv:1810.12162}, 2018.

\bibitem[Silver et~al.(2017)Silver, Hubert, Schrittwieser, Antonoglou, Lai,
  Guez, Lanctot, Sifre, Kumaran, Graepel, et~al.]{silver2017mastering}
David Silver, Thomas Hubert, Julian Schrittwieser, Ioannis Antonoglou, Matthew
  Lai, Arthur Guez, Marc Lanctot, Laurent Sifre, Dharshan Kumaran, Thore
  Graepel, et~al.
\newblock Mastering chess and shogi by self-play with a general reinforcement
  learning algorithm.
\newblock \emph{arXiv preprint arXiv:1712.01815}, 2017.

\bibitem[Still \& Precup(2012)Still and Precup]{still2012information}
Susanne Still and Doina Precup.
\newblock An information-theoretic approach to curiosity-driven reinforcement
  learning.
\newblock \emph{Theory in Biosciences}, 131\penalty0 (3):\penalty0 139--148,
  2012.

\bibitem[Sukhbaatar et~al.(2017)Sukhbaatar, Lin, Kostrikov, Synnaeve, Szlam,
  and Fergus]{sukhbaatar2017intrinsic}
Sainbayar Sukhbaatar, Zeming Lin, Ilya Kostrikov, Gabriel Synnaeve, Arthur
  Szlam, and Rob Fergus.
\newblock Intrinsic motivation and automatic curricula via asymmetric
  self-play.
\newblock \emph{arXiv preprint arXiv:1703.05407}, 2017.

\bibitem[Sun et~al.(2011)Sun, Gomez, and Schmidhuber]{sun2011planning}
Yi~Sun, Faustino Gomez, and J{\"u}rgen Schmidhuber.
\newblock Planning to be surprised: Optimal bayesian exploration in dynamic
  environments.
\newblock In \emph{International Conference on Artificial General
  Intelligence}, pp.\  41--51. Springer, 2011.

\bibitem[Torabi et~al.(2018{\natexlab{a}})Torabi, Warnell, and
  Stone]{torabi2018behavioral}
Faraz Torabi, Garrett Warnell, and Peter Stone.
\newblock Behavioral cloning from observation.
\newblock \emph{arXiv preprint arXiv:1805.01954}, 2018{\natexlab{a}}.

\bibitem[Torabi et~al.(2018{\natexlab{b}})Torabi, Warnell, and
  Stone]{torabi2018generative}
Faraz Torabi, Garrett Warnell, and Peter Stone.
\newblock Generative adversarial imitation from observation.
\newblock \emph{arXiv preprint arXiv:1807.06158}, 2018{\natexlab{b}}.

\bibitem[Tschantz et~al.(2020{\natexlab{a}})Tschantz, Baltieri, Seth, and
  Buckley]{tschantz2020scaling}
Alexander Tschantz, Manuel Baltieri, Anil~K Seth, and Christopher~L Buckley.
\newblock Scaling active inference.
\newblock In \emph{2020 International Joint Conference on Neural Networks
  (IJCNN)}, pp.\  1--8. IEEE, 2020{\natexlab{a}}.

\bibitem[Tschantz et~al.(2020{\natexlab{b}})Tschantz, Millidge, Seth, and
  Buckley]{tschantz2020reinforcement}
Alexander Tschantz, Beren Millidge, Anil~K Seth, and Christopher~L Buckley.
\newblock Reinforcement learning through active inference.
\newblock \emph{arXiv preprint arXiv:2002.12636}, 2020{\natexlab{b}}.

\bibitem[Ueltzh{\"o}ffer(2018)]{ueltzhoffer2018deep}
Kai Ueltzh{\"o}ffer.
\newblock Deep active inference.
\newblock \emph{Biological Cybernetics}, 112\penalty0 (6):\penalty0 547--573,
  2018.

\bibitem[Weihs et~al.(2019)Weihs, Kembhavi, Han, Herrasti, Kolve, Schwenk,
  Mottaghi, and Farhadi]{weihs2019artificial}
Luca Weihs, Aniruddha Kembhavi, Winson Han, Alvaro Herrasti, Eric Kolve, Dustin
  Schwenk, Roozbeh Mottaghi, and Ali Farhadi.
\newblock Artificial agents learn flexible visual representations by playing a
  hiding game.
\newblock \emph{arXiv preprint arXiv:1912.08195}, 2019.

\end{thebibliography}
\clearpage
\appendix

\section{Additional Implementation Details}
\label{sec:implementation}

\changes{\paragraph{Additional Training Details.} The experiments in the paper used two different RL algorithms for discrete action environemnts (Double DQN) and continuous action environments (TRPO). For all environments trained with Double-DQN (\Tetris, \VizDoom, \miniGrid) we use a fixed episode length of $500$ for training and collect $1000$ sample between training rounds that perform $1000$ gradient steps on the network. The replay buffer size that is used is $50000$. The same size is used for additional data buffers for RND and ICM. For \Tetris and \miniGrid network with layer sizes $[128, 64, 32]$ is used for both Q-networks. For \VizDoom the network include 3 additional convolutional layers with $[64, 32, 8]$ filters with strides $[5, 4, 3]$, all using relu activations. A learning rate of $0.003$ is used to train the Q networks.}

\changes{For the \humanoid environments the network uses relu activations with hidden layer sizes $[256, 128]$. TRPO is used to train the policy with the advantage estimated with Generalize Advantage Estimation. The training collects $4098$ sample at a time, performs $64$ gradient steps on the value function and one step with TRPO. A fixed variance is used for the policy of $0.2$ which is scaled according to the action dimensions from the environment. Each episode consisted of $4$ rounds of training and it typically take $20$ hours to train one of the \methodName policies using $8$ threads. A kl constraint of $0.2$ is used for TRPO and a learning rate of $0.001$ is used for training the value function. Next, we provide additional details on the state and action spaces of the environments and how $\theta$ was represented for each environment.}

\paragraph{Tetris}
We consider a $4\times 10$ \Tetris board with tromino shapes (composed of 3 squares).
The observation is a binary image of the current board with one pixel per square, as well as an indicator integer for the shape that will appear next. A Bernoulli distribution is used to represent the sufficient statistics $\theta$ given the to policy for \methodName. This distribution models the probability density of a block being in each of the boad locations. Double-DQN is used to train the policy for this environment. \changes{The reward function used for this environment is based on the Tetris game which gives more points for eliminating more rows at a single time.}

\paragraph{VizDoom}
For the \VizDoom environment the images are scaled down to be $48 \times 64$ grayscale. Then a history of the latest $4$ images are stacked together to use as  in separate channels. To greatly reduce the number of parameters $\theta$, \methodName needs to estimate in order to compute the state entropy the image is further reduces to $20 \time 26$. A Gaussian distribution is used to model the mean and variance over this state input. This same design is used for \VizDoomTakeCover and \VizDoomDefendTheLine. An episode timelimit of $500$ is used for each environent. Double-DQN is used to train the policy for this environment.

\textbf{Simulated \humanoid robots.}
A simulated planar \humanoid agent must avoid falling in the face of external disturbances~\citep{DBLP:journals/corr/abs-1804-06424}. %
The state-space comprises the rotation of each joint and the linear velocity of each link. 
We evaluate four versions of this task: \humanoidCliff, \humanoidTreadmill, \humanoidPedestal, and \humanoidWalk.
The \humanoidCliff task initializes the agent at the edge of a cliff, in a random pose and with a forward velocity of \valueWithUnits{$1$}{m/s}. 
Falling off the cliff leads to highly irregular and unpredictable configurations, so a surprise minimizing agent will want to learn to stay on the cliff.
In \humanoidTreadmill, the agent starts on a platform that is moving backwards at \valueWithUnits{$1$}{m/s}. %
In \humanoidPedestal, random forces are applied to it, and objects are thrown at it.
In this environment, the agent starts on a thin pedestal and random forces are applied to the robot's links and boxes of random size are thrown at the agent. 
In \humanoidWalk, we evaluate how the \methodName reward stabilizes an agent that is learning to walk.
In all four tasks, we evaluate the proportion of episodes the robot does not fall. A state is classified a fall if the agent's links, except for the feet, touch the ground, or if the agent is $-5$ meters or more below the platform or cliff. Since the state is continuous, 
We model $p_{\theta}(\bs)$ as independent Gaussian for these tasks. The full pose and link velocity state is used for the \humanoid environments $\theta$. 
The simulated robot has a control frequency of $30\texttt{hz}$. TRPO is used to train the policy for this environment.
Similar to \VizDoom $p(\bs)$ is modeled as an independent Gaussian distribution for each dimension in the observation. 
Then, the \methodName reward can be computed as:
\begin{equation*}
    r_{\text{\methodName}}(\bs) = -\sum_i \left( \log \sigma_i + \frac{(\bs_i-\mu_i)^2}{2\sigma_i^2} \right),
\end{equation*}
where $\bs$ is a single state, $ \mu_i $ and $ \sigma_i $ are calculated as the sample mean and standard deviation from $ \mathcal{D}_{t} $ and $ \bs_i $ is the $ i^{th} $ observation feature of $\bs$.

\paragraph{\miniGrid.}
This partially observed navigation environment is based on the \textit{gym\_minigrid} toolkit~\citep{gym_minigrid}. The agent vision if changed to be centered around the agent. The experiments in the paper combine \methodName with curiosity measures for \textit{Counts} that are computed using the agent locations in the discrete environment. Similar, to the \VizDoom and \humanoid environments a Gaussian distribution over the agents observations is used to estimate $\theta$. Double-DQN is used to train the policy for this environment.

\paragraph{\methodName VAE training }
The encoders and decoders of the VAEs used for \VizDoom and \humanoid experiments are implemented as fully connected networks. The coefficient for the KL-divergence term in the VAE loss was $0.1$ and $1.0$ for the \VizDoom and \humanoid experiments, respectively.
\changes{We found it very helpful to train the VAE in batches. For the \humanoid experiments where TRPO is used to train the policy the VAE is trained every 4 data collection phases for TRPO. This helped make the learning process more stationary, increasing convergence. The design of the networks used for the VAE mirrors the size and shapes of the policies used for training described earlier in this section.}

\changes{\paragraph{Fixed Length Episodes}
For \methodName it helped to used fixed length episodes during training to help keep \methodName from terminating early. For example, in the \VizDoom environments \methodName would result in policies that would terminate as soon as possible so the agent would return to a similar initial state. In fact, for training we need to turn on god mode to prevent this behaviour. Similarly, to discourage SMiRL from terminating Tetris early by quickly stacking pieces in the same tower (resulting in low entropy) we added "soft resets" where the simulation will reset when the game fails and the episode will continue on forcing the SMiRL agent to learn how to eliminate rows to reduce the number of blocks in the scene.}

\section{\methodName Distributions}
\label{sec:smirl_distributions}

\paragraph{\methodName on \Tetris.} In \Tetris, since the state is a binary image, we model $p(\bs)$ as a product of independent Bernoulli distributions for each board location. The \methodName reward $\log p_{\theta}(\bs)$ becomes:
\begin{equation*}
    r_{\text{\methodName}}(\bs) = \sum_i \bs_i \log \theta_i + (1 - \bs_i) \log (1 - \theta_i ),
\end{equation*}
where $\bs$ is a single state, the update procedure $\theta_i=\mathcal U(\mathcal{D}_{t})$ returns the sample mean of $\mathcal{D}_t$, indicating the proportion of datapoints where location $ i $ has been occupied by a block, and $s_i$ is a binary variable indicating the presence of a block at location $ i $.
If the blocks stack to the top, the game board resets, but the episode continues and the dataset $\data_{t}$ continues to accumulate states.

\newlength{\tetrisw}
\setlength{\tetrisw}{0.11\linewidth}
\begin{figure*}[t]
	\centering
	\includegraphics[trim={1.75cm 0.0cm 0.0cm 8.0cm},clip,width=\tetrisw]{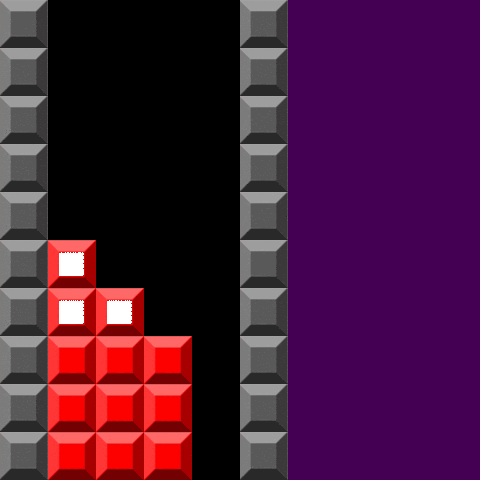} 
	\includegraphics[trim={1.75cm 0.0cm 0.0cm 8.0cm},clip,width=\tetrisw]{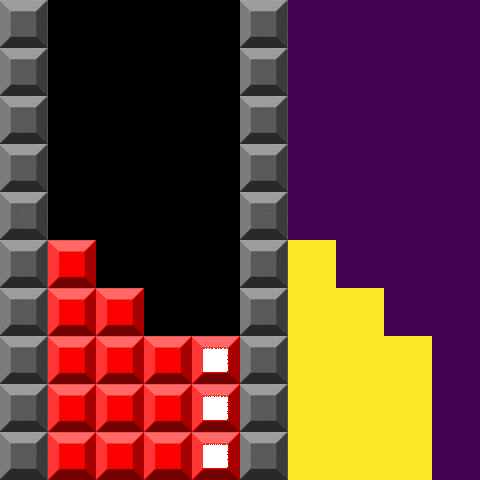} 
	\includegraphics[trim={1.75cm 0.0cm 0.0cm 8.0cm},clip,width=\tetrisw]{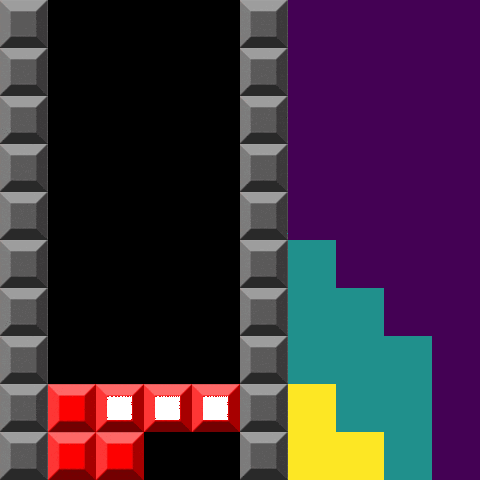} 
	\includegraphics[trim={1.75cm 0.0cm 0.0cm 8.0cm},clip,width=\tetrisw]{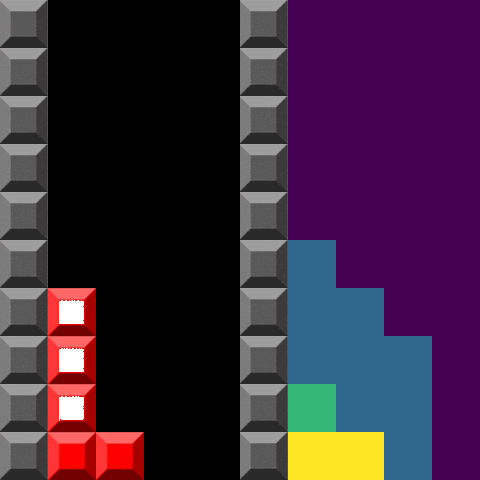} 
	\includegraphics[trim={1.75cm 0.0cm 0.0cm 8.0cm},clip,width=\tetrisw]{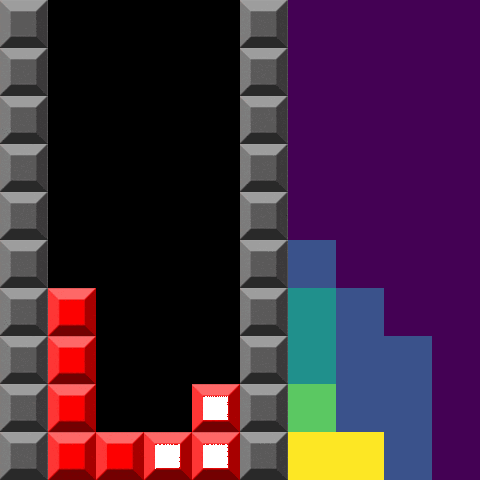} 
	\includegraphics[trim={1.75cm 0.0cm 0.0cm 8.0cm},clip,width=\tetrisw]{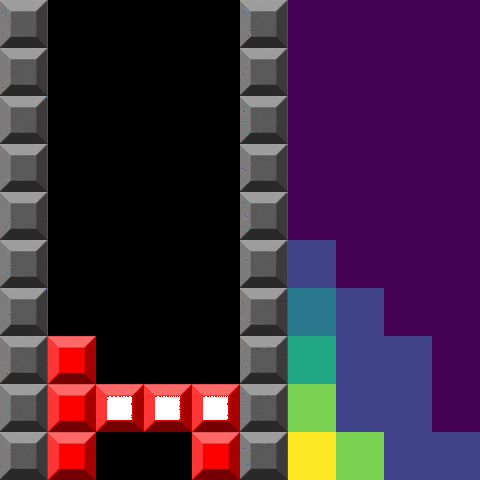} 
	\includegraphics[trim={1.75cm 0.0cm 0.0cm 8.0cm},clip,width=\tetrisw]{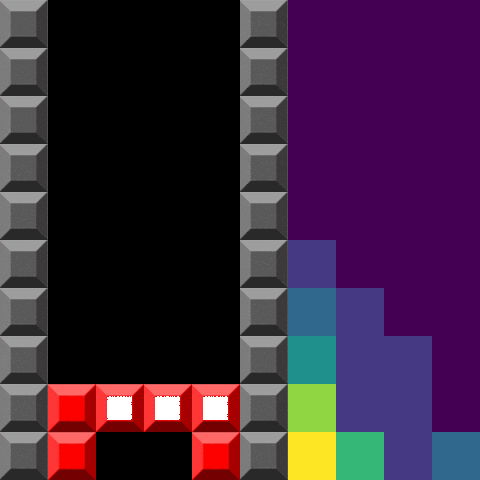} 
	\includegraphics[trim={1.75cm 0.0cm 0.0cm 8.0cm},clip,width=\tetrisw]{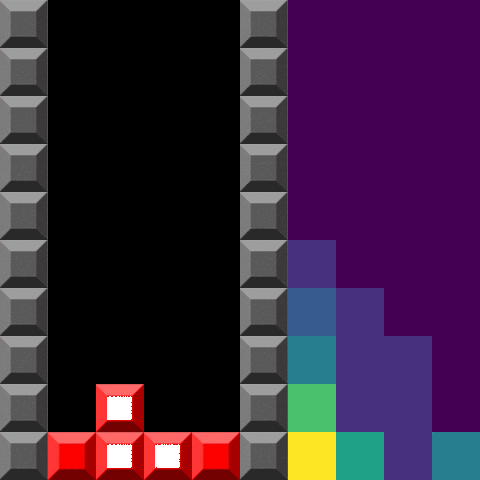} \\
	
	\includegraphics[trim={1.75cm 0.0cm 0.0cm 8.0cm},clip,width=\tetrisw]{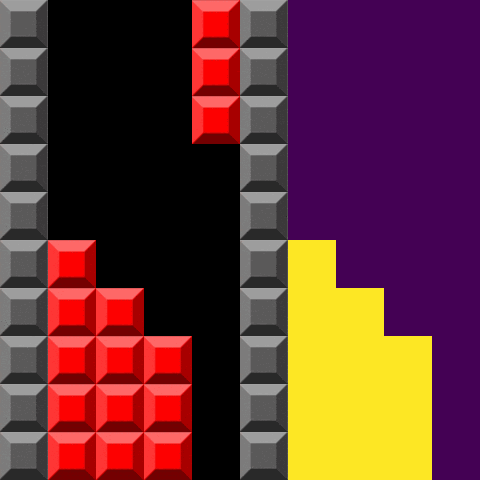}
	\includegraphics[trim={1.75cm 0.0cm 0.0cm 8.0cm},clip,width=\tetrisw]{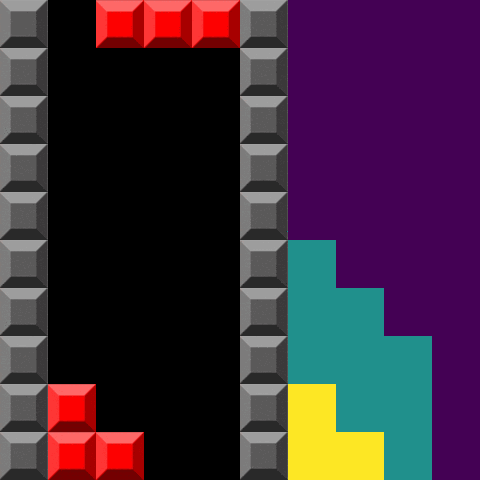}
	\includegraphics[trim={1.75cm 0.0cm 0.0cm 8.0cm},clip,width=\tetrisw]{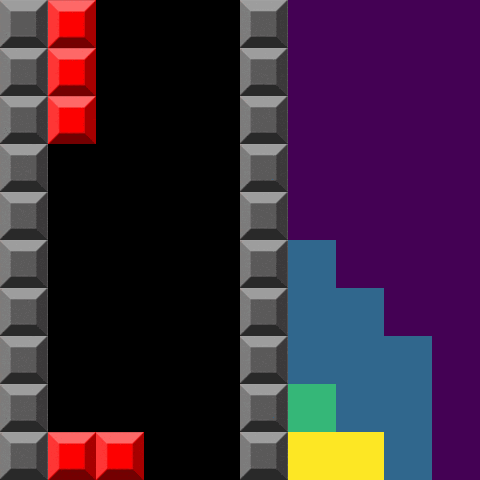}
	\includegraphics[trim={1.75cm 0.0cm 0.0cm 8.0cm},clip,width=\tetrisw]{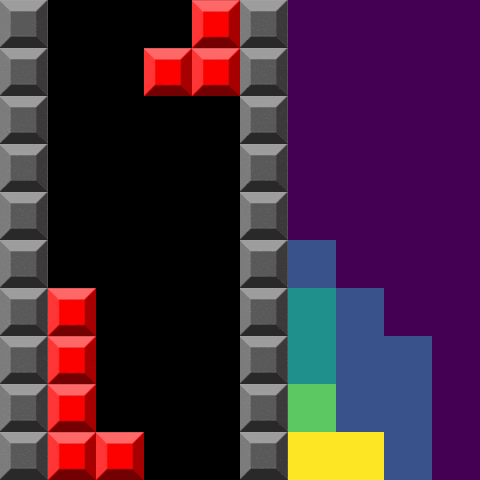}
	\includegraphics[trim={1.75cm 0.0cm 0.0cm 8.0cm},clip,width=\tetrisw]{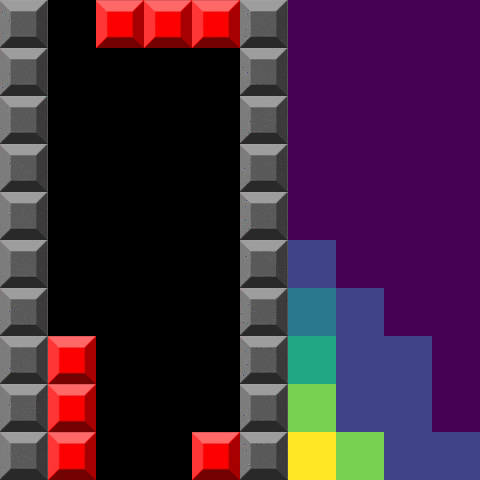}
	\includegraphics[trim={1.75cm 0.0cm 0.0cm 8.0cm},clip,width=\tetrisw]{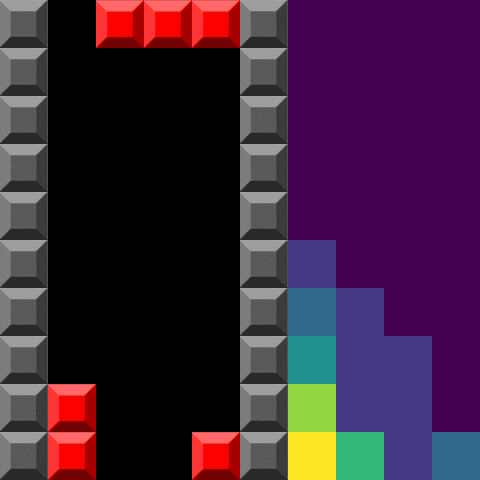}
	\includegraphics[trim={1.75cm 0.0cm 0.0cm 8.0cm},clip,width=\tetrisw]{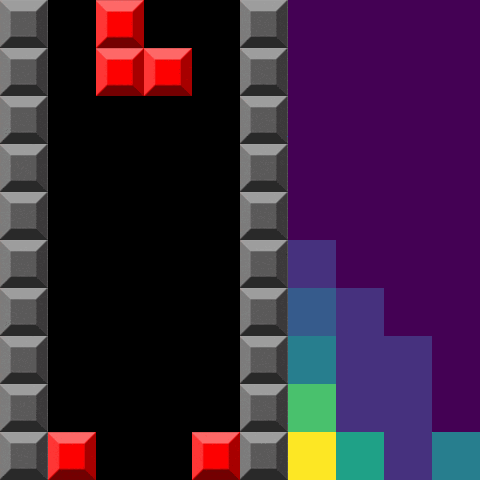}
	\includegraphics[trim={1.75cm 0.0cm 0.0cm 8.0cm},clip,width=\tetrisw]{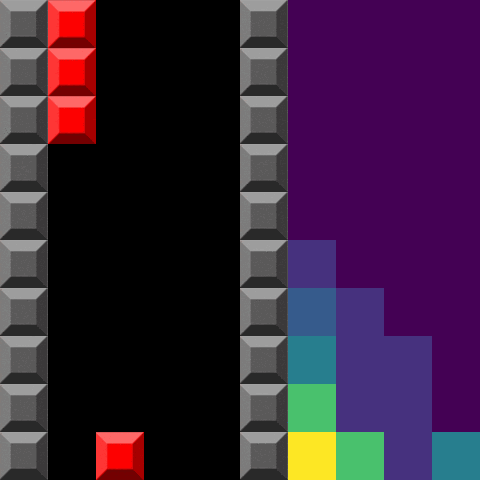}
	\caption{\label{fig:smirl_rollout} Frames from \Tetris, with state $\bs$ on the left and parameters $\theta_{t}$ of an independent Bernoulli distribution for each board location on the right, with higher probability shown in yellow. The top row indicates the newly added block and bottom row shows how the state changes due to the newly added block along with the updated $\theta_{t}$.}
\end{figure*}

\section{\methodName MDP} 
\label{app:smirl_MDP}

Note that the RL algorithm in \methodName is provided with a standard stationary MDP (except in the VAE setting, more on that below), where the state is augmented with the parameters of the belief over states $\theta$ and the timestep $t$. We emphasize that this MDP is Markovian, and therefore it is reasonable to expect any convergent \RL algorithm to converge to a near-optimal solution. Consider the augmented state transition $p(s_{t+1}, \theta_{t+1}, t+1 | s_{t}, a_{t}, \theta_{t}, t )$. This transition model does not change over time because the updates to $\theta$ are deterministic when given $s_t$ and $t$. The reward function $r(s_{t}, \theta_{t}, t)$ is also stationary, and is in fact deterministic given $ s_{t}$ and $\theta_{t}$. Because \methodName uses \RL in an MDP, we benefit from the same convergence properties as other \RL methods.

\paragraph{Transition dynamics of $\theta_t$.}
\label{app:smirl_dynamics}
Given the augmented state $(\bs_t, \theta_t, t)$, we show that the transition dynamics of the MDP are Markovian.
The $\bs_t$ portion of the augmented state are from the environment, therefore all convergence properties of RL hold. 
Here we show that $(\theta_t, t)$ is also Markovian given $\bs_{t+1}$. To this end, we describe the transition dynamics of $(\theta_t, t)$ for an incremental estimation of a Gaussian distribution, which is used in most experiments. Here we outline $\theta_{t+1} = \mathcal U(\bs_t, \theta_t, t)$.
\begin{align*}
    \theta_t &= (\mu_{t}, \sigma_{t}^{2}) \\
    \mu_{t+1} &= \frac{t\mu_{t} + \bs_t}{t+1} \\
    \sigma_{t+1}^{2} &= \frac{t(\sigma_{t}^{2} + \mu_{t}^{2}) + \bs_t}{t+1} - \mu_{t+1}^{2} \\
    \theta_{t+1} &= (\mu_{t+1}, \sigma_{t+1}^{2}) \\
    t_{t+1} &= t_{t} + 1 \\
\end{align*}
These dynamics are dependant on the current augmented state $(\bs_t, \theta_t, t)$ and the next state $\bs_{t+1}$
of the RL environment and do not require an independent
model fitting process.

However, the version of \methodName that uses a representation learned from a VAE is not Markovian due to not adding the VAE parameters to the state $\bs$, and thus the reward function changes over time. We find that this does not hurt results, and note that many intrinsic reward methods such as ICM and RND also lack stationary reward functions.

\section{More Environment Stability Details}
\label{sec:full_stability_env_analysis}
Here we include the full data on the stability analysis in~\refFigure{tab:entropy-compare-table-full}. From this data and the additional results on the \href{https://sites.google.com/view/surpriseminimization}{website}  we can see the \methodName can reduce the entropy of a few of the Atari environments as well. These include Assault, where \methodName hides on the left but is good at shooting ships and Carnival, where \methodName also reduces the number of moving objects. RND on the other hand tends to induce entropy and cause many game flashes.

\begin{table}
\vspace{-0.25cm}
    \centering
    \begin{tabular}{ccccc}
    \toprule
      Environment   & RND & Random & \methodName & Relative \\ 
      \midrule
        \textit{Tetris} & 18.6$\pm 2.7$ & 17.1$\pm 1.8$ & 5.2$\pm 2.1$ & -43.4 \\ 
        \VizDoomTakeCover & -4.7$\pm 0.7$ & -5.9$\pm 1.1$ & -13.2$\pm 0.7$ & -10.4 \\ 
        \VizDoomDefendTheLine & 19.6$\pm 0.6$ & 19.9$\pm 0.7$ & -23.4$\pm 0.4$ & -8.5\\
        \hline
        \textit{Assault} & 193.1$\pm 1.4$ & 181.8$\pm 2.7$ & 124.9$\pm 2.3$ & -45.6 \\
        \textit{SpaceInvaders} & 208.4$\pm 3.4$ & 206.5$\pm 5.2$ & 196.3$\pm 4.2$ & -8.3\\
        \textit{Carnival} & 151.2$\pm 1.4$ & 130.8$\pm 2.7$ & 107.7$\pm 4.3$ & -2.7\\
        \textit{RiverRaid} & 264.4$\pm 3.4$ & 269.1$\pm 2.2$ & 274.9$\pm 3.2$ & 0.3\\
        \textit{Gravitar} & 198.6$\pm 1.7$ & 167.8$\pm 2.7$ & 141.3$\pm 1.3$ & 4.3\\ 
        \textit{Berzerk} & 197.2$\pm 1.4$ & 180.0$\pm 2.7$ & 177.1$\pm 4.7$ & 14.3\\ 
        \bottomrule
    \end{tabular}
    \caption{Estimated entropies for three of our tasks, and an example Atari games studied by \citet{burda2018rnd}, where novelty-seeking exploration works well. Note the large \emph{negative} \textit{Relative} entropy gap in our tasks with overall lower initial entropy, which are both absent in most Atari games. \changes{This data shows the mean and std over 3 seeds.}}
    \label{tab:entropy-compare-table-full}
\vspace{-0.5cm}
\end{table}

\section{Addition Notes on Unsupervised RL Related Work}

\changes{
The works in ~\citet{tschantz2020reinforcement,annabi2020autonomous} are interesting and discuss connections to active inference and RL. However, these methods and many based on active inference “encode” the task reward function as a “global prior” and minimizing a KL between the agents state distribution this “global prior”. Our work instead actively estimates a marginal over the distribution of states the agent visits (with no prior data) and then minimizes this “online” estimate of the marginal, as is described in Section 3. Our work differs from LP-based methods~\citep{kim2020active,lopes2012exploration,schmidhuber1991curious} because SMiRL is learning to control the marginal state distribution rather than identifying the system parameters.}

\section{Additional Results}

\begin{figure}%
\vspace{-0.0cm}
\centering
  \includegraphics[trim={0.0cm 0.0cm 0.0cm 0.0cm},clip,width=0.32\linewidth]{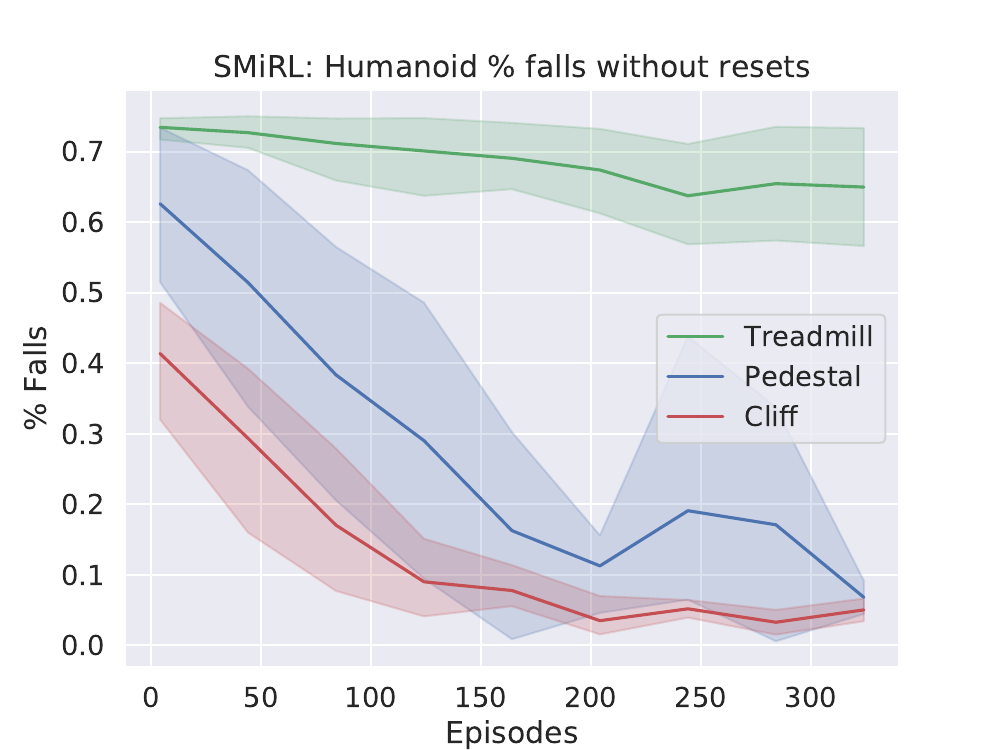}
\caption{ \methodName results when the \humanoid environments are trained without early termination based resets (fixed episode lengths). \humanoidCliff and \humanoidPedestal still produce entropy minimizing policies that reduce falls. RL has difficulty with optimizing the more challenging \humanoidTreadmill environment.
}
\label{fig:biped_results2}
\vspace{-0.5cm}
\end{figure}

To better understand the types of behaviors \methodName produces we conducted an experiment with fixed episode lengths on the \humanoid environments~(\refFigure{fig:biped_results2}). This shows that \methodName results in surprise minimizing behaviors independent of how long the episode is.

\end{document}